\documentclass[11pt]{article}

\usepackage[margin=1in]{geometry}

\usepackage{microtype}
\usepackage{graphicx}
\usepackage{subfigure}
\usepackage{booktabs}
\usepackage{hyperref}
\usepackage{url}

\usepackage{amsmath}
\usepackage{amssymb}
\usepackage{mathtools}
\usepackage{multirow}
\usepackage{amsthm}
\usepackage{bm}

\usepackage[capitalize,noabbrev]{cleveref}

\usepackage{enumitem}

\usepackage{authblk}

\theoremstyle{plain}
\newtheorem{theorem}{Theorem}[section]

\newtheorem{lemma}[theorem]{Lemma}

\theoremstyle{definition}

\theoremstyle{remark}

\newcommand{\R}{\mathbb{R}}

\newcommand{\Tr}{\operatorname{Tr}}

\usepackage{natbib}
\begin{document}
\title{Rod Flow: A Continuous-Time Model for Gradient Descent \\ at the Edge of Stability}
\author{Eric Regis$^{*}$ \quad Sinho Chewi$^{*}$}
\date{}
\maketitle
\renewcommand{\thefootnote}{}
\footnotetext{$^{*}$Yale University. Email: \texttt{\{eric.regis, sinho.chewi\}@yale.edu}}
\renewcommand{\thefootnote}{\arabic{footnote}}
\begin{abstract}
How can we understand gradient-based training over non-convex landscapes?
The edge of stability phenomenon, introduced in \citet{cohen2021gradient}, indicates that the answer is not so simple: namely, gradient descent (GD) with large step sizes often diverges away from the gradient flow.
In this regime, the ``Central Flow'', recently proposed in \citet{Cohen+25CentralFlow}, provides an accurate ODE approximation to the GD dynamics over many architectures. In this work, we propose \emph{Rod Flow}, an alternative ODE approximation, which carries the following advantages: (1) it rests on a principled derivation stemming from a physical picture of GD iterates as an extended one-dimensional object---a ``rod''; (2) it better captures GD dynamics for simple toy examples and matches the accuracy of Central Flow for representative neural network architectures, and (3) is explicit and cheap to compute.
Theoretically, we prove that Rod Flow correctly predicts the critical sharpness threshold and explains self-stabilization in quartic potentials. We validate our theory with a range of numerical experiments.
\end{abstract}


\section{Introduction}
\label{sec:introduction}
Machine learning models are trained by performing optimization on non-convex loss landscapes. Traditional analysis in convex optimization assumes that the loss landscape is well-behaved---specifically, that the largest eigenvalue of the Hessian (the \emph{sharpness}) is bounded by $2/\eta$ where $\eta$ is the learning rate. This assumption underlies classical convergence guarantees via the descent lemma.

However, this assumption is routinely violated in practice. When training neural networks with full-batch gradient descent, \citet{cohen2021gradient} observed that the sharpness does not remain bounded below $2/\eta$. Instead, it \emph{increases} during training (\emph{progressive sharpening}) until it reaches $2/\eta$, and then hovers at this threshold (\emph{edge of stability}). This phenomenon is remarkably robust, occurring across architectures, datasets, and learning rates.

The edge of stability poses a fundamental challenge for continuous-time approximations. The classical approach to understanding gradient descent is to study gradient flow, the ODE $\dot{w} = -\nabla L(w)$. But approximating gradient descent by gradient flow requires slowly-varying trajectories, while gradient descent at the edge of stability exhibits rapid oscillations along the sharpest direction of the Hessian.

Several works have attempted to model edge of stability dynamics \citep{arora2022understanding, DamNicLee23SelfStab, Cohen+25CentralFlow}, but existing approaches either lack formal justification or fail to accurately track the discrete iterates. Central Flow \citep{Cohen+25CentralFlow} is a prior continuous-time model designed to capture the time-averaged trajectory of oscillatory optimizers. It tracks the ``center'' of oscillations $\bar{w}$ along with a covariance matrix $\Sigma$ constrained by a semidefinite complementarity problem. However, its derivation relies on heuristic approximations.

We introduce \textbf{Rod Flow}, a system of ODEs that models gradient descent as an extended one-dimensional object---a ``rod''---characterized by its center $\bar{w}$ and extent $\Sigma$. The physical picture is intuitive: rather than tracking a point that oscillates rapidly, we track an extended object whose endpoints sample the loss landscape at $\bar{w} + \delta$ and $\bar{w} - \delta$. The key insight is that, while individual iterates $w_t$ oscillate rapidly, the \emph{average} of consecutive iterates $\bar{w}_t$ and the \emph{outer product} of the half-difference of consecutive iterates $\delta_t \otimes \delta_t$  both evolve gradually and are amenable to ODE approximation.

\begin{figure}[t]
\centering
\includegraphics[width=0.48\textwidth]{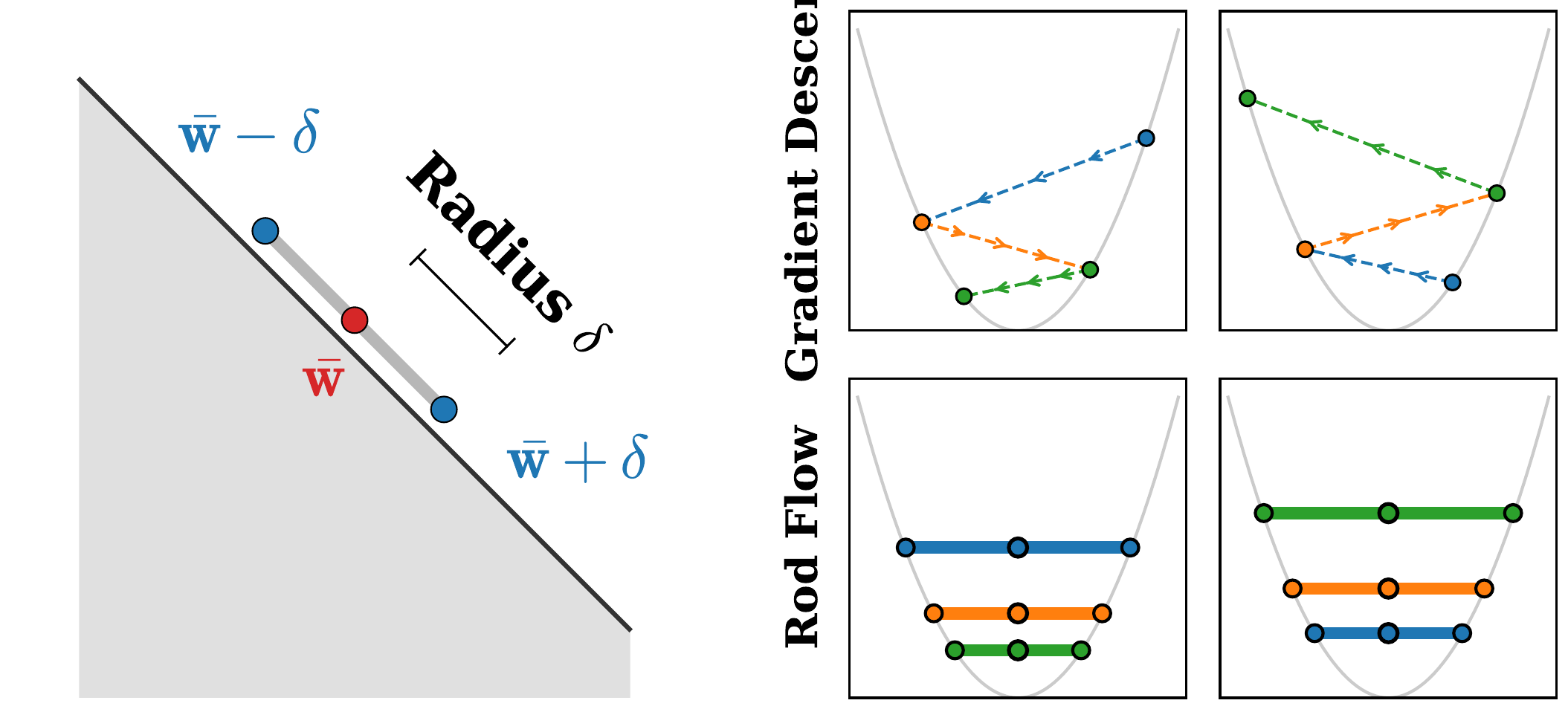}
\hfill
\includegraphics[width=0.50\textwidth]{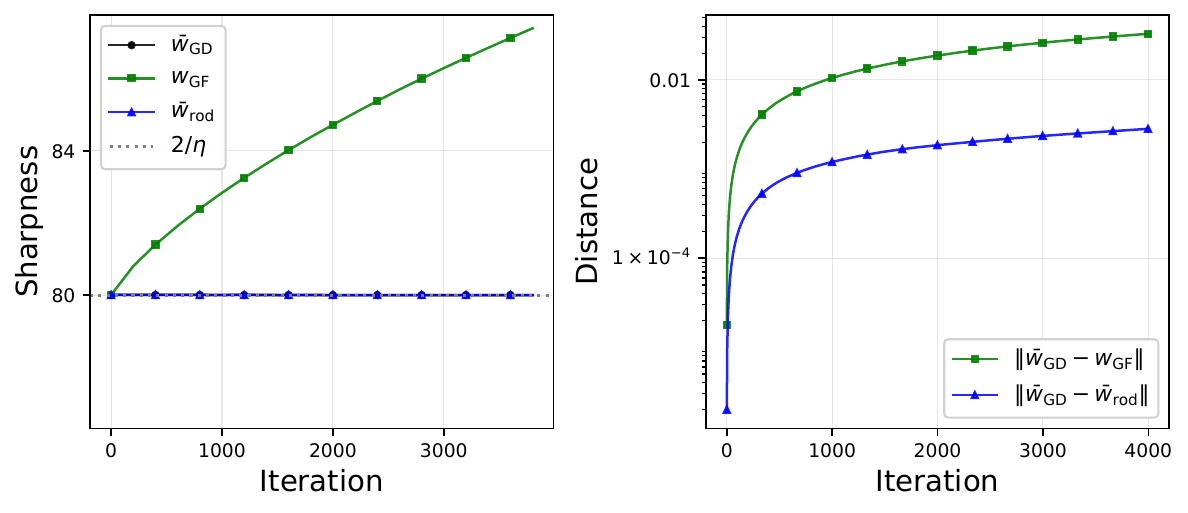}
\caption{ \textbf{Illustration of Rod Flow.} \textbf{Left:} A rod moving down a flat loss landscape, where the two endpoints represent consecutive GD iterates and the midpoint represents their average. \textbf{Center Left:} Comparison of gradient descent (top row) and Rod Flow (bottom row) on a quadratic loss landscape under convergent (left column) and divergent (right column) step sizes. \textbf{Right:} Comparison of gradient descent, gradient flow, and Rod Flow on a 3-layer MLP. The left plot shows sharpness over time: gradient descent and rod flow remain at the edge of stability while gradient flow continues to sharpen. The right plot shows the distance between the GD iterates and the two flows, demonstrating that Rod Flow stays closer to the GD trajectory than gradient flow.}
\label{fig:combined}
\end{figure}
\subsection{Contributions} Our main contributions are as follows:

\begin{itemize}[leftmargin=*,itemsep=0pt,topsep=2pt]
    \item A principled derivation of Rod Flow from discrete dynamics via exact difference equations and backward error analysis (\cref{sec:rodflow})

    \item A theoretical analysis showing how Rod Flow correctly captures the $2/\eta$ stability threshold and exhibits self-stabilization in quartic potentials (\cref{sec:theory})

    \item  A detailed comparison of Rod Flow with Central Flow, showing how Central Flow arises as a Taylor approximation to Rod Flow (\cref{sec:comparison})

    \item An empirical demonstration that Rod Flow exhibits high accuracy on toy problems and comparable accuracy to Central Flow across representative deep learning architectures (\cref{sec:experiments})

\end{itemize}


 \section{Related Work}
  \label{sec:related}

  \textbf{Edge of stability.} \citet{cohen2021gradient} documented progressive sharpening and edge of stability across
  architectures, observing that sharpness rises to $2/\eta$ and then hovers. Theoretical explanations include
  self-stabilization mechanisms \citep{DamNicLee23SelfStab}, two-step analysis \citep{ahn2023learning, chen2023beyond},
  convergence guarantees in the unstable regime \citep{ahn2022understanding, arora2022understanding}, and bifurcation
  theory \citep{song2023trajectory}. \citet{ahn2023learning, zhu2023understanding} provide minimalist examples, while
  \citet{AgaPedPen23SecondOrder} show progressive sharpening in quadratic regression. The catapult mechanism
  \citep{lewkowycz2020large} explains how large learning rates enable escape from sharp initializations.
  Additional related work includes loss landscape analysis \citep{MaKunYin22BeyondQuadratic}, balancing effects with large learning rates \citep{Wang+23EoS}, chaotic dynamics in gradient descent \citep{Che+24StabToChaos}, and NTK evolution at the edge of stability \citep{JiaCohLi25NTKEoS}.

  \textbf{Continuous-time models.} Gradient flow is the classical continuous limit of gradient descent but fails at edge
   of stability since it cannot capture oscillations. \citet{rosca2023continuous} study instabilities in gradient flow
  models. The closest work to ours is \citet{Cohen+25CentralFlow}, which introduces the Central Flow that tracks a covariance matrix $\Sigma$ constrained
  by a semidefinite complementarity problem (SDCP) in the subspace of critical eigenvectors. Our Rod Flow provides a
  simpler and more principled alternative derived from backward error analysis.

\textbf{Backward error analysis.} Modified equations from backward error analysis are standard in numerical analysis
  \citep{wilkinson1963rounding, wilkinson1965algebraic, hairer2006geometric}. \citet{barrett2021implicit} and \citet{smith2021origin} use this framework to understand
  implicit regularization in gradient descent. \citet{shi2021continuous} derive high-resolution ODEs for momentum
  methods, while \citet{li2017stochastic} develop stochastic modified equations for SGD.
  \citet{digiovacchino2024backward} provide methodological validation for applying backward error analysis to
  optimization algorithms. We extend this framework to derive ODEs that capture edge of stability dynamics.


 \section{Gradient Descent Dynamics}\label{sec:gradient-descent}

  We consider minimizing a loss function $L: \mathbb{R}^p \to \mathbb{R}$ using gradient descent with learning rate
  $\eta$:
  \begin{equation}
      w_{t+1} = w_t - \eta \nabla L(w_t)\,.
  \end{equation}
  Classical convergence analysis relies on the \emph{descent lemma}, which guarantees progress when the loss is
  sufficiently smooth.

  \begin{lemma}[Descent Lemma]
  If $L$ has $S$-Lipschitz gradients ($\|\nabla^2 L\| \leq S$ everywhere), then for $\eta < 2/S$:
  \begin{equation}
      L(w_{t+1}) \leq L(w_t) - \eta\, \Bigl(1 - \frac{\eta S}{2}\Bigr)\, \|\nabla L(w_t)\|^2\,.
  \end{equation}
  \end{lemma}
  When the sharpness (largest eigenvalue of the Hessian) exceeds $2/\eta$, this guarantee breaks down and gradient
  descent may diverge---indeed, this occurs on quadratic objectives. However, the empirical behavior of neural network training is more nuanced. We identify three distinct phases of training (Figure \ref{fig:edge_of_stability}):

\begin{figure}[t]
\centering
\includegraphics[width=0.60\textwidth]{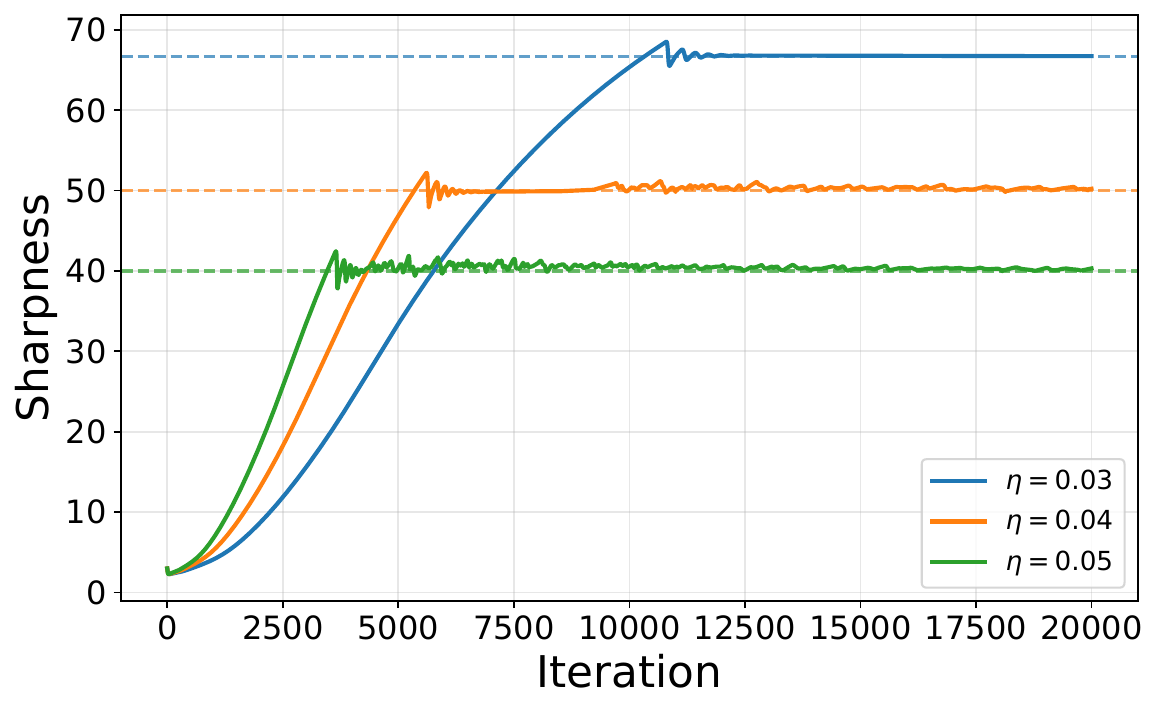} \hspace{2em}
\includegraphics[width=0.30\textwidth]{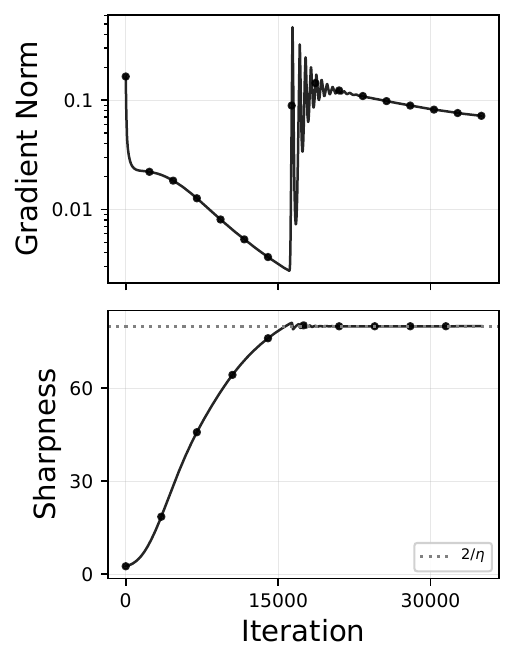}  \hspace{4em}
\caption{\textbf{Edge of Stability.} \textbf{Left:} Edge of stability exhibited for three step sizes. For all step sizes, we observe progressive sharpening followed by stabilization of the sharpness at $2/\eta$. \textbf{Right:} The three phases of edge of stability. Phase I (steady descent): progressive sharpening with decreasing gradient norms. Phase II (entering edge of stability): gradient norm spikes and fluctuates wildly. Phase III (steady-state edge of stability): sharpness remains at $2/\eta$ while gradient norm becomes smooth.}
\label{fig:edge_of_stability}
\end{figure}

 \textbf{Phase 1: Stable Descent.} Early in training, the sharpness remains below $2/\eta$. The loss decreases monotonically, and gradient flow provides an accurate continuous-time approximation of gradient descent. This phase is characterized by \emph{progressive sharpening}: the sharpness steadily increases toward the critical threshold \citep{cohen2021gradient, AgaPedPen23SecondOrder}.

\textbf{Phase 2: Transition to Edge of Stability.} When the sharpness first reaches $2/\eta$, training enters a turbulent transitory phase characterized by large spikes and fluctuations in the norm of the gradient, with concomitant fluctuations in the loss and sharpness. These events share the self-stabilization mechanism identified by \citet{lewkowycz2020large} in their study of ``catapults''. Unlike the single large catapult that occurs when training is \emph{initialized}, here progressive sharpening \emph{drives} the sharpness above the threshold repeatedly, producing multiple transient instabilities before the dynamics settle.

\textbf{Phase 3: Steady-State Edge of Stability.} After the transitory phase, training settles into a regime where the sharpness hovers at $2/\eta$ but the oscillation magnitude varies smoothly without dramatic spikes \citep{cohen2021gradient, DamNicLee23SelfStab}. The loss continues to decrease on average despite local non-monotonicity, and the dynamics exhibit regular, bounded oscillations. The transient instabilities of Phase 2 involve rapid, large-amplitude changes in the gradient that are difficult to capture with any continuous-time approximation. However, the steady-state oscillations of Phase 3 are smooth and regular, making them amenable to continuous-time modeling. This is precisely what Rod Flow aims to achieve: a principled ODE approximation that accurately tracks gradient descent dynamics during the steady-state edge of stability regime.

\section{Rod Flow}
\label{sec:rodflow}
\subsection{Setup and Notation}
We define the \emph{center} and \emph{half-difference} of consecutive iterates:
\begin{equation}
\bar{w}_t = \frac{w_{t+1} + w_t}{2}\,, \quad \delta_t = \frac{w_{t+1} - w_t}{2}\,.
\end{equation}
From these definitions, we recover the original iterates as $w_t = \bar{w}_t - \delta_t$ and $w_{t+1} = \bar{w}_t + \delta_t$. Using the gradient descent update rule $w_{t+1} = w_t - \eta \nabla L(w_t)$, we have $\delta_t = -\frac{\eta}{2} \nabla L(w_t)$.

Let $u \otimes v \coloneqq uv^\top$ denote the outer product. The quantity $\delta_t \otimes \delta_t$ captures the magnitude and direction of the half-difference. Its continuous-time counterpart is denoted $\Sigma$. We use $\Sigma$ rather than outer product notation because $\Sigma$ is only approximately rank-1 as it evolves (and it also notationally aligns with Central Flow's covariance matrix). Empirically, $\Sigma$ remains approximately rank-1 with ratios of $10^2$--$10^3$ between its largest and second largest eigenvalues. So we identify $\delta$ with the principal eigenvector of $\Sigma$ scaled by the square root of its principal eigenvalue. For brevity, we write $L_+ \coloneqq L(\bar{w} + \delta)$ and $L_- \coloneqq L(\bar{w} - \delta)$, with $\nabla L_\pm$ and $\nabla^2 L_\pm$ denoting the corresponding gradients and Hessians.

\subsection{The Rod Flow Equations}

Rod Flow is defined by a system of ODEs for the center $\bar{w}$ and the extent $\Sigma$:
\begin{align}
    \frac{d\bar{w}}{dt} &= -\frac{\eta}{2}\big[\nabla L_+ + \nabla L_-\big] \notag \\
    &\quad - \frac{\eta^2}{8}\big[\nabla^2 L_+ + \nabla^2 L_-\big]\big[\nabla L_+ + \nabla L_-\big]\,, \label{eq:wbar_ode} \\
    \frac{d\Sigma}{dt} &= \frac{\eta^2}{4}\big[\nabla L_+ \otimes \nabla L_+ + \nabla L_- \otimes \nabla L_-\big] - 2\Sigma\,. \label{eq:sigma_ode}
\end{align}
The center evolves in accordance with the average of the forces experienced at the two endpoints $\bar{w} \pm \delta$. The extent grows based on the gradient magnitudes at the endpoints, with a restoring force biasing $\Sigma$ toward zero.

The remainder of this section is devoted to deriving these two equations. We first derive exact difference equations for $\bar{w}_t$ and $\delta_t \otimes \delta_t$, then promote them to ODEs via backward error analysis.

\subsection{Exact Difference Equations}
\label{sec:difference_eqs}

\subsubsection{Difference Equation for \texorpdfstring{$\bar{w}_t$}{the Center}}

For the center $\bar{w}_t$, we compute:
\begin{align}
    \bar{w}_{t+1} - \bar{w}_t &= \frac{w_{t+2} + w_{t+1}}{2} - \frac{w_{t+1} + w_t}{2} \nonumber \\
    &= \frac{w_{t+2} - w_{t+1}}{2} + \frac{w_{t+1} - w_t}{2} \nonumber \\
    &= -\frac{\eta}{2}\big[\nabla L_+ + \nabla L_-\big]\,. \label{eq:wbar_diff}
\end{align}
The final line follows from using the gradient descent update rule when applied to both $w_{t+1}$ and $w_{t}$. We then invoked the fact that $w_{t+1} = \bar{w}_t + \delta_t$ and $w_{t} = \bar{w}_t - \delta_t$.

\subsubsection{Difference Equation for \texorpdfstring{$\delta_t \otimes \delta_t$}{the Extent}}

To derive the difference equation for $\delta_t \otimes \delta_t$, we first add and subtract $\delta_t \otimes \delta_t$:
\begin{equation}
\delta_{t+1} \otimes \delta_{t+1} - \delta_t \otimes \delta_t
= \delta_{t+1} \otimes \delta_{t+1} + \delta_t \otimes \delta_t - 2\delta_t \otimes \delta_t\,.
\end{equation}
Using $\delta_t = -\frac{\eta}{2}\nabla L(w_t)$, we then have that:
\begin{align}
    &\delta_{t+1} \otimes \delta_{t+1} - \delta_t \otimes \delta_t \nonumber \\
    &\quad = \frac{\eta^2}{4}\big[\nabla L_+ \otimes \nabla L_+ + \nabla L_- \otimes \nabla L_-\big] - 2\delta_t \otimes \delta_t\,. \label{eq:delta_diff}
\end{align}
\subsubsection{Why Track $\delta \otimes \delta$ Instead of $\delta$?}
\label{sec:why_outer_product}

\begin{figure}[t]
\begin{center}
\centerline{\includegraphics[width=\textwidth]{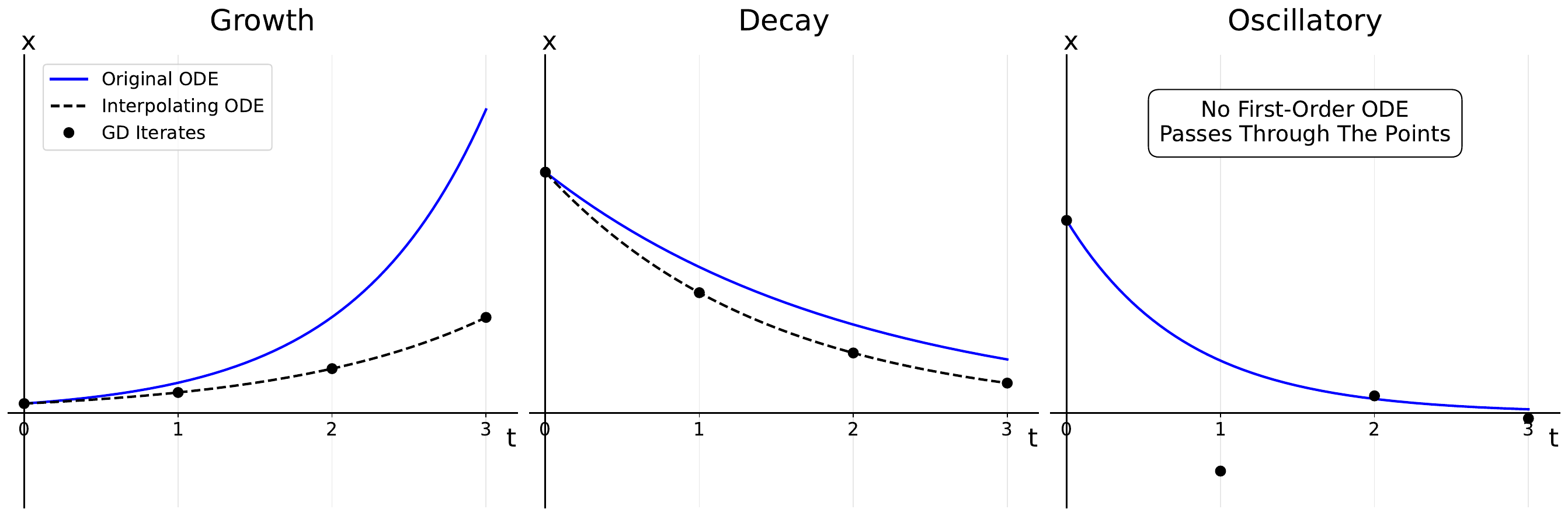}}
\caption{\textbf{Backward Error Analysis.} The ODE that passes through discrete points differs from the ODE whose discretization generates them. \textbf{Left:} Exponential growth. \textbf{Middle:} Exponential decay. \textbf{Right:} GD iterates alternate in sign and no first-order ODE passes through them.}
\label{fig:ode_gd}
\end{center}
\end{figure}

A natural question arises: why track $\delta_t \otimes \delta_t$ rather than $\delta_t$ directly?

The answer lies in the behavior at the edge of stability. When sharpness exceeds $2/\eta$, the GD iterates oscillate along the direction of the sharp axis, causing the corresponding component of $\delta_t$ to alternate signs. A first-order ODE cannot capture oscillations. Any ODE for $\delta_t$ would therefore fail to capture edge of stability.

However, even when $\delta_t$ oscillates, its \emph{magnitude} $\|\delta_t\|$ can vary smoothly. Consider the extreme case of when $\delta_{t+1} = -\delta_t$. Even though $\delta_t$ flips its sign every time step, $\delta_t \otimes \delta_t$ remains constant.

\subsubsection{Properties of the Difference Equations}

\textbf{Both \cref{eq:wbar_diff} and \cref{eq:delta_diff} are exact}---no approximations have been made. We have simply rewritten the gradient descent recurrence in different variables.

Two properties deserve emphasis. First, both equations are symmetric under $\delta_t \mapsto -\delta_t$, which is essential because under our formalism the sign of $\delta_t$ is arbitrary. Second, both equations depend only on $(\bar{w}_t, \delta_t)$, forming a closed system independent of the original iterates.

\subsection{Backward Error Analysis}
\label{sec:bea}

We now promote the difference equations to ODEs. The key insight is that the ODE whose Euler discretization defines the recurrence relation $x_{t+1} - x_t = V(x_t)$ is \emph{not} the ODE that passes through the discrete-time iterates---discretization error must be accounted for.

\subsubsection{A Motivating Example}

Consider a concrete example. The ODE $\dot{x} = kx$ has the solution $x(t) = x_0 e^{kt}$. Discretizing with step size $\Delta t = 1$ gives:
\begin{equation}
    x_{t+1} - x_t = k x_t \implies x_{t+1} = (1+k)x_t\,.
\end{equation}
The iterates satisfy $x_t = (1+k)^t x_0 = e^{\ln(1+k) \cdot t} x_0$. So the ODE passing through the discrete points is $\dot{x} = \ln(1+k) x$, not $\dot{x} = kx$ (\cref{fig:ode_gd}).

\subsubsection{General Backward Error Analysis}

Given a difference equation $x_{t+\Delta t} - x_t = \Delta t \ V(x_t)$, we seek an ODE whose solutions interpolate the discrete points. Assuming a smooth curve $x(t)$ passes through the iterates, Taylor expansion gives:
\begin{equation}
    x(t+\Delta t) = x(t) + \Delta t \ \dot{x}(t) + \frac{\Delta t^2}{2}\ddot{x}(t) + O(\Delta t^3).
\end{equation}
Matching the above with $x(t+\Delta t) = x(t) +  \Delta t \ V(x(t))$ and using $\ddot{x} = \nabla V(x) \, \dot{x}$ for autonomous ODEs, we obtain:
\begin{equation}
    \dot{x} = V(x) - \frac{\Delta t}{2}\,\nabla V(x) \cdot V(x) + O(\Delta t^2)\,. \label{eq:bea}
\end{equation}
The first term on the right-hand side is the na\"{\i}ve continuous limit. We must go to order $\Delta t$ to get the first correction offered by backward error analysis. We obtain the modified vector field $\tilde{V}(x) = V(x) - \frac{1}{2}\nabla V \cdot V$. A more detailed exposition on backward error analysis appears in \cref{app:bea}.

\subsubsection{Application to $\bar{w}$}

\cref{eq:wbar_diff} has the form $\bar{w}_{t+1} - \bar{w}_t = V(\bar{w}_t, \delta_t)$ where:
\begin{equation}
V(\bar{w}, \delta) = -\frac{\eta}{2}\big[\nabla L_+ + \nabla L_-\big]\,.
\end{equation}
Applying \cref{eq:bea}, the Jacobian is:
\begin{equation}
\nabla_{\bar{w}} V = -\frac{\eta}{2}\big[\nabla^2 L_+ + \nabla^2 L_-\big]\,.
\end{equation}
The backward-error-corrected ODE is therefore $\frac{d\bar{w}}{dt} = V - \frac{1}{2}\nabla_{\bar{w}} V \cdot V$, which yields \cref{eq:wbar_ode}.

\subsubsection{Application to $\Sigma$}

For the $\Sigma$ equation, we omit the backward error correction. The rationale is time-scale separation: the $-2\Sigma$ decay term in \cref{eq:delta_diff} operates at rate 2, while $\bar{w}$ changes at rate $O(\eta)$. When $\eta \ll 1$, this separation means $\Sigma$ rapidly equilibrates to a quasi-steady-state determined by the current $\bar{w}$. Since $\Sigma$ remains near its instantaneous equilibrium, the backward error correction has negligible effect. We therefore promote \cref{eq:delta_diff} directly to \cref{eq:sigma_ode}.


\section{Theoretical Analysis}
\label{sec:theory}

The simple and explicit structure of Rod Flow allows us to analyze its behavior on canonical loss functions and to build intuition. Our goal is to understand the fixed points of the $(\bar{w}, \Sigma)$ system and their stability properties.

\subsection{Flat Landscape}

Consider the linear loss $L(w) = -b \cdot w$ with constant gradient $\nabla L = -b$ and zero curvature. The Rod Flow equations become:
\begin{align}
    \frac{d\bar{w}}{dt} &= \eta b\,, \\
    \frac{d\Sigma}{dt} &= \frac{\eta^2}{2}b \otimes b - 2\Sigma\,.
\end{align}

The center moves at constant velocity $\eta b$. Setting $\frac{d\Sigma}{dt} = 0$ gives the steady-state extent:
\begin{equation}
    \Sigma^* = \frac{\eta^2}{4}b \otimes b\,.
\end{equation}

This matches the discrete dynamics exactly: consecutive GD iterates are separated by $\eta\|b\|$, so the half-difference has magnitude $\|\delta_t\| = \frac{\eta}{2}\|b\|$.

\subsection{Quadratic Loss}

Consider the quadratic $L(w) = \frac{1}{2}Sw^2$ with $S > 0$. For simplicity, we will work in one dimension. As $\Sigma$ is a scalar, we have that $\delta = \sqrt{\Sigma}$.

By inspection, $(\bar{w}^*, \Sigma^*) = (0, 0)$ is a fixed point of Rod Flow. We will analyze its stability by considering initial conditions: $\bar{w}(0) = 0$ and $\Sigma(0) > 0$. The Rod Flow equations become:
\begin{align}
    \frac{d\bar{w}}{dt} &= 0\,, \label{eq:quad_wbar} \\
    \frac{d\Sigma}{dt} &= \Bigl(\frac{\eta^2 S^2}{2} - 2\Bigr)\Sigma\,. \label{eq:quad_sigma}
\end{align}
The center remains stationary. For $\Sigma$, define $\beta = \frac{\eta^2 S^2}{4} -1$. Then:
\begin{equation}
    \frac{d\Sigma}{dt} = 2\beta\Sigma \implies \Sigma(t) = \Sigma(0)e^{2\beta t}\,.
\end{equation}
The extent decays when $\beta < 0$ and grows when $\beta >0$. In order for the $(0,0)$ fixed point to be stable, we must have:
\begin{equation}
    \beta < 0 \implies  S < \frac{2}{\eta}\,.
\end{equation}
This recovers the edge of stability threshold: the fixed point $(0, 0)$ is stable when the sharpness $S < 2/\eta$, and unstable when $S > 2/\eta$.

\subsection{Quartic Potential}

The quadratic analysis predicts unbounded growth in $\Sigma$ when $S > 2/\eta$. In practice, oscillations stabilize at finite amplitude. To capture this, we include higher-order terms.

Consider an even quartic potential in one dimension:
\begin{equation}
    L(w) = \frac{S}{2}w^2 -\frac{Q}{4}w^4\,,
\end{equation}
where $S > 0$ is the curvature at the origin and $Q > 0$ controls the quartic term. We have $\nabla L(w) = Sw - Qw^3$.

At $\bar{w} = 0$ and with $\delta = \sqrt{\Sigma}$, we have that:
\begin{equation}
    \nabla L(\pm\delta) = \pm(S\sqrt{\Sigma} - Q\Sigma^{3/2})\,.
\end{equation}
Once again, we have that $\frac{d\bar{w}}{dt} = 0$. For the ODE for $\Sigma$, we have that:
\begin{align}
   \frac{d \Sigma}{dt} &= \Bigl(\frac{\eta^2 S^2}{2} - 2\Bigr)\Sigma - \eta^2 SQ\Sigma^2 + \frac{\eta^2 Q^2}{2}\Sigma^3\,. \label{eq:quartic_sigma_main}
\end{align}
We find steady-state solutions for $\Sigma$ by setting $\frac{d\Sigma}{dt} = 0$. One steady-state solution is $\Sigma^* = 0$. The non-zero solutions satisfy the quadratic:
\begin{equation}
    \frac{\eta^2 Q^2}{2}(\Sigma^{*})^2 - \eta^2 SQ\Sigma^* + \frac{\eta^2 S^2}{2} - 2 = 0\,. \label{eq:quartic_quadratic_main}
\end{equation}
Using the quadratic formula, we can solve for the non-zero roots:
\begin{equation}\Sigma^* = \frac{S}{Q} \pm \frac{2}{\eta Q}\,.\end{equation}
The root $\Sigma^{*} = S/Q + 2/\eta Q$ is unstable, corresponding to the runaway region of the quartic potential. The root of interest is $\Sigma^* = S/Q - 2/\eta Q$. Since only positive values of $\Sigma$ are well-defined, we require:
$$\frac{S}{Q} - \frac{2}{\eta Q} > 0 \implies S > \frac{2}{\eta}\,.$$

When $S > 2/\eta$, this solution is stable.

This analysis provides a mechanistic explanation for the empirical observations of \citet{cohen2021gradient}. During training, as iterates approach a minimum, the sharpness $S$ tends to increase---this is progressive sharpening. When $S < 2/\eta$, the rod contracts: iterates converge toward the minimum, and the sharpness increases. When $S > 2/\eta$, the rod expands: iterates diverge along the top eigenvector of the Hessian, but this expansion samples regions where higher-order terms in the loss landscape reduce the effective curvature. This is \textit{self-stabilization} \citep{DamNicLee23SelfStab}: the nonlinearity of the loss function naturally restores stability by decreasing sharpness whenever iterates diverge due to instability.

\subsection{Fixed Points and Stability}

We now discuss fixed points of Rod Flow more generally. A fixed point is a pair $(\bar{w}^*, \Sigma^*)$ satisfying $\frac{d\bar{w}}{dt} = 0$ and $\frac{d\Sigma}{dt} = 0$.

Let $S(\bar{w}^*)$ denote the sharpness at $\bar{w}^*$. There are two types of fixed points when $\bar{w}^*$ is a critical point of the loss ($\nabla L(\bar{w}^*) = 0$):

\textbf{Quiescent fixed points:} $\Sigma^* = 0$. These are stable when $S(\bar{w}^*) < 2/\eta$. Gradient descent converges to $\bar{w}^*$ without sustained oscillations.

\textbf{Edge-of-stability fixed points:} $\Sigma^* > 0$. These arise when $S(\bar{w}^*) > 2/\eta$ and higher-order terms in the loss provide a stabilizing effect (as in the quartic example). The extent $\Sigma^*$ reflects the amplitude of sustained oscillations around $\bar{w}^*$.

If $\bar{w}^*$ is a critical point of the loss, then $(\bar{w}^*, 0)$ is always a fixed point of Rod Flow. The key difference between gradient flow and Rod Flow is that, with Rod Flow, \emph{the stability of the $\Sigma = 0$ fixed point depends on the sharpness at $\bar{w}^*$}. A more detailed fixed point analysis is performed in Appendix \ref{app:fixedpoints}.

\section{Comparison with Central Flow}
\label{sec:comparison}

\begin{figure}[t]
\centering
\includegraphics[width=\textwidth]{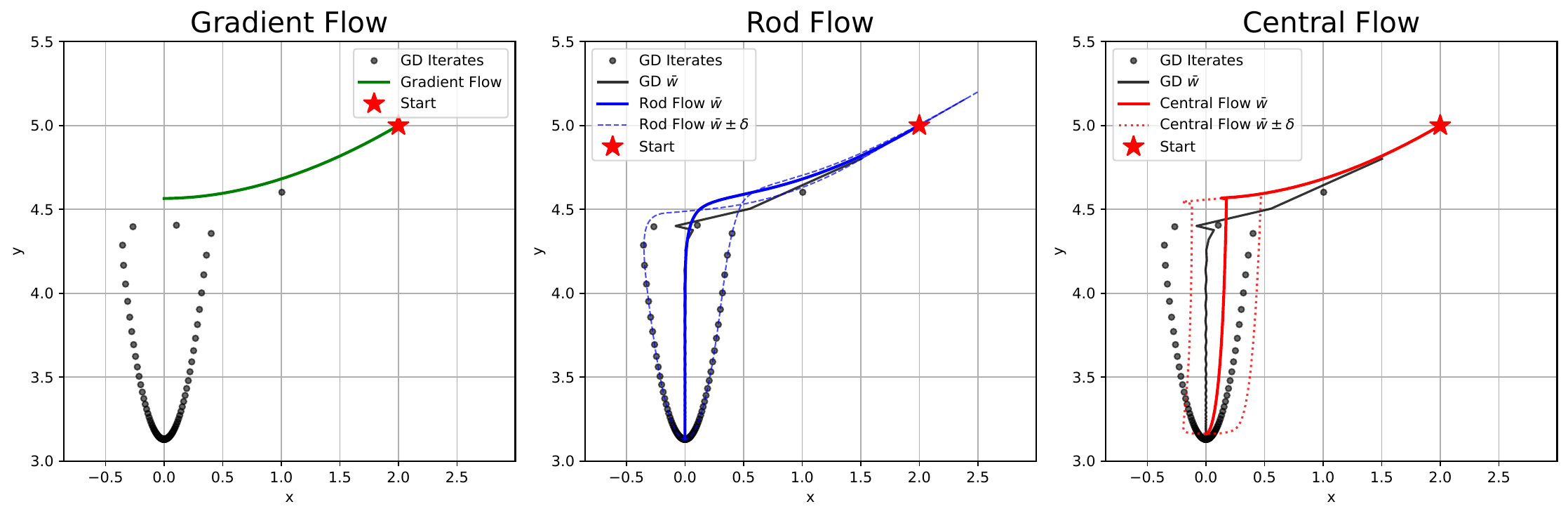}
\caption{\textbf{Comparison on 2D square root loss} $L(x,y) = \sqrt{1 + (xy)^2}$. This loss has minima along the axes, with sharpness increasing away from the origin. When the local minimum is too sharp, GD bounces until the sharpness decreases to $2/\eta$. Gradient flow converges to the local minimum irrespective of the sharpness. Rod Flow tracks GD precisely. Central Flow follows the sharpness manifold.}
\label{fig:2d_comparison}
\end{figure}

Central Flow \citep{Cohen+25CentralFlow} is a prior continuous-time model for edge of stability. We compare the two approaches.

 \subsection{Central Flow Overview}

  The core insight of Central Flow is that while exact oscillatory dynamics are difficult to analyze, the
  \emph{time-averaged} trajectory is tractable. Central Flow derives an ODE for this smoothed trajectory, viewing
  $\bar{w}$ as the local average of oscillating iterates. It defines a matrix $\Sigma$ which represents the covariance of the oscillations of the GD iterates during edge of stability. $\Sigma$ is restricted to the subspace of \emph{critical eigenvectors}---those with eigenvalue near $2/\eta$.

 The center evolves according to:
  \begin{equation}
      \frac{d\bar{w}}{dt} = -\eta \nabla L(\bar{w}) - \frac{\eta}{2}\Tr(\Sigma \nabla^3 L)\,,
  \end{equation}
  where the first term is gradient flow and the second accounts for how oscillations bias motion toward regions of lower
   curvature.

  The covariance $\Sigma$ is also dynamic, but unlike $\bar{w}$, its evolution is \emph{not} given by an ODE. Instead,
  $\Sigma$ is determined at each instant by solving a semidefinite complementarity problem (SDCP).

Central Flow also admits a geometric interpretation. It can be viewed as \emph{projected gradient flow}: the dynamics follow the gradient but are constrained to the feasible region $\{\bar{w} : S(\bar{w}) \le 2/\eta\}$ with $\Sigma$ serving as the Lagrange multiplier enforcing this constraint. More details on Central Flow are provided in \cref{app:centralflow}.

\subsection{Key Differences}

\begin{figure}[t]
    \centering
    \includegraphics[width=\textwidth]{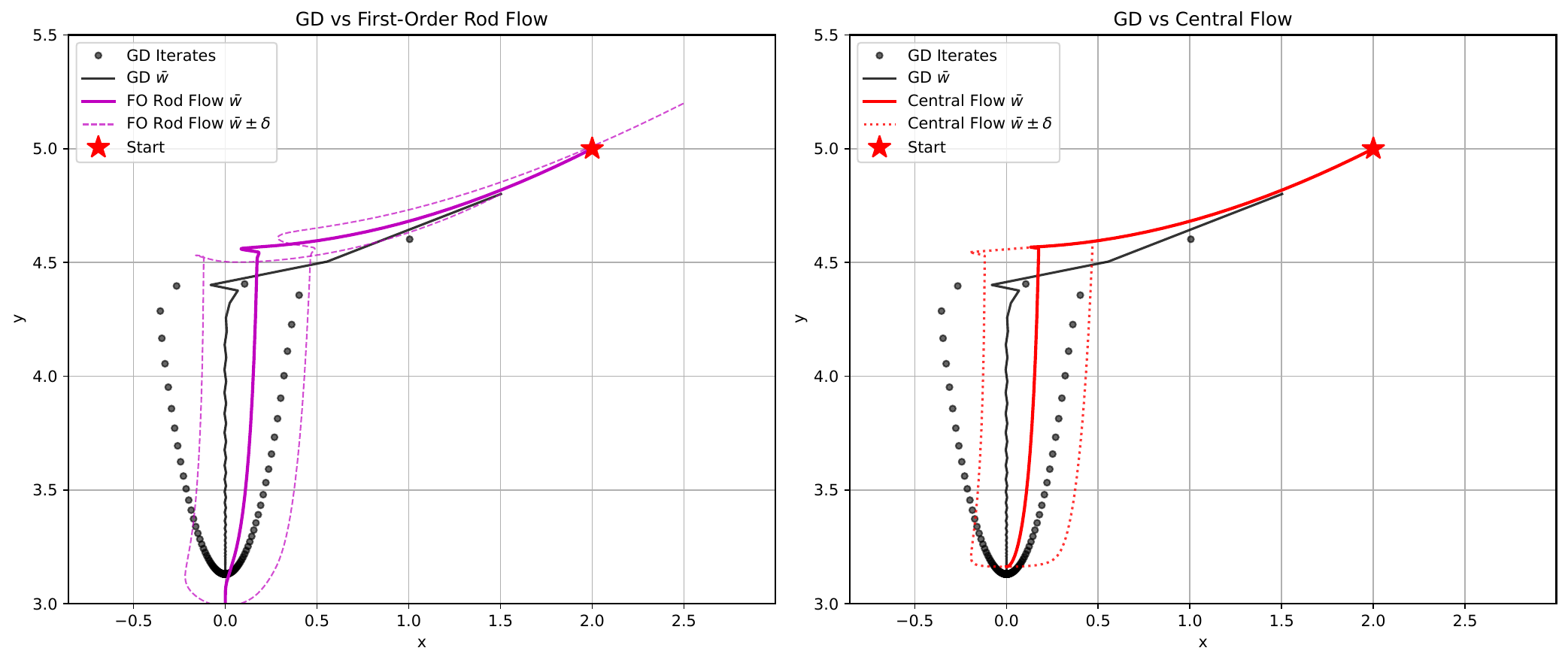}
    \caption{\textbf{Comparison of First-Order Rod Flow and Central Flow.}}
    \label{fig:fo-rod-vs-central}
\end{figure}

We briefly cover some of the key differences between Rod Flow and Central Flow.

\textbf{Derivation.} Central Flow's derivation involves Taylor expansions in the critical subspace and Lagrange multipliers for the stability constraint. Rod Flow is derived via exact difference equations and standard backward error analysis.

\textbf{Simplicity.} Central Flow's dynamics are complex due to $\Sigma$ being determined by an SDCP. In contrast, Rod Flow yields a simpler ODE that is more amenable to mathematical analysis (as we demonstrated in \cref{sec:theory}).

\textbf{Physical interpretation.} Central Flow interprets the covariance as Gaussian spread around $\bar{w}$. Rod Flow offers a more literal picture: the rod has endpoints at $\bar{w} \pm \delta$, which correspond to the actual locations of the iterates.

\textbf{Evaluation points.} Central Flow evaluates gradients at the center $\bar{w}$ and uses Taylor expansion to account for the displacement.  Rod Flow evaluates gradients at the rod endpoints $\bar{w} \pm \delta$. This makes Rod Flow's non-locality a feature: it directly samples the loss landscape where the iterates actually visit.

\textbf{Computational cost.} Central Flow requires  solving an SDCP---a complex optimization problem---at each timestep. Rod Flow uses only gradient evaluations and Hessian-vector products, both $O(p)$ operations. The $\Sigma$ matrix for Rod Flow is stored in low-rank form (rank 3 suffices). More information about the computational implementation of Rod Flow can be found in \cref{app:computation}.

\subsection{Connection Between the Flows}

Central Flow can be viewed as a Taylor approximation of Rod Flow. Expanding $\nabla L_{\pm}$ to second order in $\delta$:

\begin{align}
    \nabla L(\bar{w} \pm \delta) &\approx \nabla L(\bar{w}) \pm \nabla^2 L(\bar{w})\delta + \frac{1}{2}\nabla^3 L(\bar w)[\delta, \delta]\,.
\end{align}

Consider the ODE for $\bar{w}$. If we drop the backward error analysis term (due to being of order $\eta^2$) and we substitute the above expression for $\nabla L_\pm$, we have that:

\begin{equation}
    \frac{d\bar{w}}{dt} \approx -\eta \nabla L(\bar{w}) - \frac{\eta}{2}\nabla^3 L(\bar w)[\delta, \delta]\,.
\end{equation}
If $\delta$ aligns with the top eigenvector of $\nabla^2 L$, then $\nabla^3 L(\bar w)[\delta, \delta] \approx \|\delta\|^2 \nabla S$, recovering Central Flow's ODE for $\bar{w}$.

By expanding the ODE for $\Sigma$ to second order in $\delta$, we can derive a simplified ODE for $\Sigma$ as well. In this expansion, we will neglect the gradient term and retain only the term involving the Hessian:
\begin{equation}
\frac{d\Sigma}{dt} \approx \frac{\eta^2}{2} (\nabla^2 L \cdot \delta)(\nabla^2 L \cdot \delta)^\top - 2 \Sigma
\end{equation}

We dub the resulting system of ODEs ``first-order Rod Flow,'' as the equations are expanded to first order in $\Sigma$. On toy examples, the trajectory of first-order Rod Flow closely matches that of Central Flow, though the timescales of the two flows differ. This is depicted in \cref{fig:fo-rod-vs-central}.


\section{Experiments}
\label{sec:experiments}

\begin{figure}[t]
\centering
\includegraphics[width=0.32\textwidth]{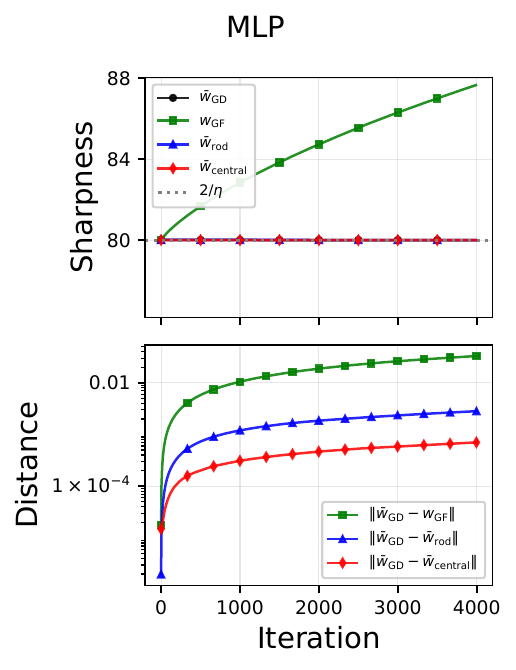}
\hfill
\includegraphics[width=0.32\textwidth]{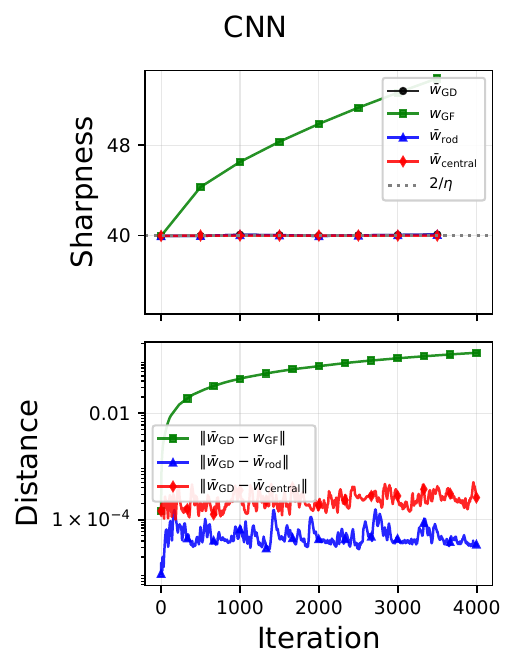}
\hfill
\includegraphics[width=0.32\textwidth]{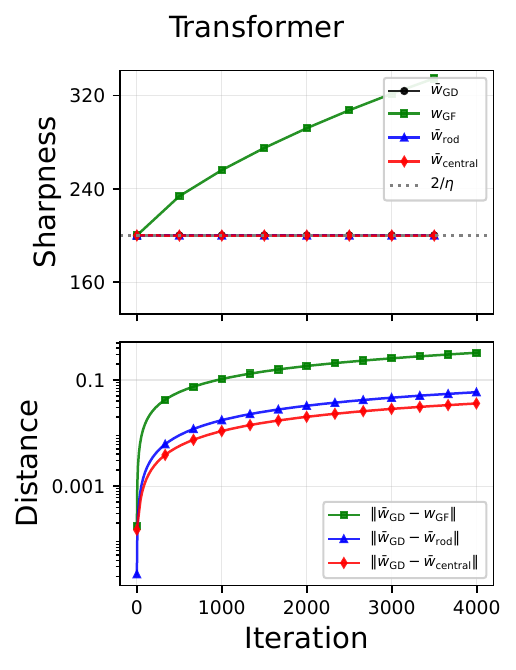}
\caption{\textbf{Experimental Results on Neural Networks.}  Comparison of gradient flow, Rod Flow, and Central Flow on three different architectures (MLP, CNN, and Transformer). The three flows were initialized after a warm-up phase by gradient descent. Gradient flow continues to sharpen while Rod Flow and Central Flow exhibit self-stabilization. Both Rod Flow and Central Flow are orders of magnitude more accurate in terms of distance to GD than gradient flow. }
\label{fig:centers}
\end{figure}

We compare gradient descent (GD), gradient flow (GF), Rod Flow (RF), and Central Flow (CF) across both toy problems and representative neural network architectures.

\subsection{Square Root Loss (2D)}

We consider the following two-dimensional loss, which was studied in~\citet{ahn2023learning} as a minimal example of edge of stability (\cref{fig:2d_comparison}).
\begin{equation}
    L(x, y) = \sqrt{1 + (x y)^2}.
\end{equation}
This loss has minima along the $x$ and $y$ axes (where $xy = 0$), with the sharpness of the minima increasing with distance from the origin along each axis.

Depending on the initial conditions, the GD iterates can exhibit edge of stability. If the local minimum is too sharp, the iterates will bounce along the perpendicular axis, decreasing the sharpness until they can converge to a local minimum with sharpness around $2/\eta$.

When tested on the square root loss, the three flows exhibit different behaviors. Gradient flow converges to the nearest local minimum, irrespective of how sharp it is. Rod Flow tracks the GD iterates precisely: $\bar{w}_{\text{RF}}$ quickly converges along the $y$-axis, and the envelope $\bar{w}_{\text{RF}} \pm \delta_{\text{RF}}$ closely matches the location of the actual GD iterates. Central Flow follows the sharpness manifold, causing $\bar{w}_{\text{CF}}$ to lie between the actual iterates and the average of the iterates.

\subsection{Neural Networks}

To further compare the different flows, we performed experiments on various neural network architectures: a 3-layer MLP, a 3-layer CNN, and a 2-layer Transformer.

Due to turbulence at the onset of edge of stability, we ran gradient descent until the iterates were safely ensconced in the steady-state edge of stability regime (Phase III; see \cref{sec:gradient-descent}). We then initialized all three flows (GF, RF, and CF) at the location of the GD iterates after this warm-up period. The results are shown in \cref{fig:centers}.

Both Rod Flow and Central Flow clearly outperform gradient flow. While gradient flow exhibits increasing sharpness, Rod Flow and Central Flow both exhibit self-stabilization, with their respective sharpnesses remaining at the critical value of $2/\eta$. Additionally, both CF and RF stay closer to the actual GD trajectory as measured by the distance between their respective $\bar{w}$ and the true average of the iterates.

Rod Flow shows stronger performance than Central Flow on the CNN while performing somewhat worse on the MLP and Transformer. These variations in performance may be attributable to architecture-specific differences in the spectrum of the Hessian. More experimental details are provided in \cref{app:neuralnetworks}.


\section{Conclusion}

We introduced Rod Flow, a continuous-time model for gradient descent that accurately captures dynamics at the edge of stability. By modeling gradient descent as an extended object---a rod with center $\bar{w}$ and half-length $\delta$---we obtain a system of ODEs that tracks both the average position and the half-difference of the iterates.

Our derivation via exact difference equations and backward error analysis provides a principled foundation. Theoretically, Rod Flow correctly predicts the $2/\eta$ stability threshold and exhibits self-stabilization in quartic potentials. Empirically, Rod Flow is better at capturing edge of stability phenomena than gradient flow, and it performs on par to Central Flow while being simpler to implement.

\textbf{Limitations.} Rod Flow, like all ODE approximations of gradient descent, assumes that the dynamics are sufficiently smooth. During edge of stability in large neural networks, the gradient norm occasionally spikes by orders of magnitude, causing discrete jumps that no ODE can capture. Characterizing the loss function regularity conditions under which Rod Flow remains accurate is an open problem. Additionally, while Rod Flow is more computationally efficient than alternative continuous-time models, it is still expensive relative to gradient descent, making it impractical as a substitute for actual training runs. Finally, the mechanism behind progressive sharpening---why the sharpness increases to $2/\eta$ in the first place---remains unexplained.

\textbf{Future outlook.} We hope that the idea of modeling discrete-time algorithms using extended objects will help shed light toward a general theory of non-convex optimization---beyond just the setting of deep learning.

\section*{Acknowledgements}

We thank the authors of the Central Flow paper for making their code publicly available. Our implementation builds on code from \url{https://github.com/alex-damian/EOS} and \url{https://github.com/centralflows/centralflows}.


\bibliography{references}


\newpage
\appendix

\section{Backward Error Analysis}
\label{app:bea}

\subsection{Historical Context}

We provide a brief overview of backward error analysis, a technique from numerical analysis that allows us to understand discrete-time algorithms through the lens of continuous dynamics.

The term "backward error analysis" originates in numerical linear algebra, largely through the work of Wilkinson in the 1960s \citep{wilkinson1963rounding, wilkinson1965algebraic}. The classical setting considers computing an approximate solution $\tilde{y}$ to a problem $y = f(x)$. \emph{Forward error analysis} asks: how far is the computed answer from the true answer? This involves bounding the quantity $\|\tilde{y} - f(x)\|$. \emph{Backward error analysis} inverts the question: can we find a perturbation to the input such that the computed answer is \emph{exactly} correct? This involves finding $\Delta x$ such that $\tilde{y} = f(x + \Delta x)$. The power of the backward error analysis perspective is that if the backward error $\|\Delta x\|$ is small and the problem is well-conditioned, we can conclude the forward error must also be small.

A related but distinct usage of "backward error analysis" arose in the study of numerical integrators for ODEs, particularly in the geometric integration literature. Here the question becomes: given a numerical method (say, Euler's method) that does not exactly solve $\dot{x} = V(x)$, is there a \emph{modified} ODE $\dot{x} = V(x) + \Delta t \, V_1(x) + \Delta t^2 V_2(x) + \cdots$ that the numerical iterates \emph{do} lie on? The logic parallels Wilkinson's approach: the vector field is the input and the flow map is the output. Instead of viewing discretization error as a forward error (the iterates deviate from the true flow), we view the iterates as following a perturbed system exactly. We work ``backward'' from the discrete dynamics to find the continuous system they actually solve.

This is the sense in which we use backward error analysis to derive Rod Flow.

\subsection{Backward Error Analysis on the Quadratic}

A natural question arises: why do we apply backward error analysis to the transformed variable $\bar{w}$ rather than directly to the original iterates $w$? The answer lies in the radius of convergence of the perturbative expansion for the modified vector field $\tilde{V}$: it is precisely in the edge of stability regime that this expansion breaks down. We can see this most clearly in the case of a quadratic loss, where we can solve for the modified vector field to all orders in $\eta$ exactly.

Consider the quadratic loss $L(w) = \frac{1}{2}Sw^2$. The gradient flow ODE is:

$$\frac{dw}{dt} = - \eta S w,$$

where we use the convention that one unit of time corresponds to one gradient descent step. The gradient descent recurrence is:

\begin{equation}
    w_{t+1} = w_t - \eta S w_t = (1 - \eta S) w_t.
\end{equation}
We seek an ODE $\dot{w} = \tilde{V}(w)$ whose flow passes through the discrete iterates.

\paragraph{The easy approach.} There is an elegant way to find $\tilde{V}$ without resorting to perturbation theory. After $t$ units of time, we have:
\begin{equation}
    w_t = (1 - \eta S)^{t} w_0 = e^{t \log(1 - \eta S)} w_0.
\end{equation}
By inspection, this is the solution to the ODE:
\begin{equation}
    \dot{w} = \log(1 - \eta S) \, w.
\end{equation}
Taylor expanding the rate in $\eta S$ gives:
\begin{equation}
    \log(1 - \eta S) = -\eta S - \frac{(\eta S)^2}{2} - \frac{(\eta S)^3}{3} - \cdots
\end{equation}
The leading term $-\eta S \, w = -\eta \nabla L(w)$ recovers gradient flow, with higher-order corrections in $\eta$.

\paragraph{The perturbative approach.} It is instructive to derive the modified vector field using the full backward error analysis machinery, as this approach generalizes beyond quadratics. We assume there exists a first-order ODE $\dot{w} = \tilde{V}(w)$ passing through the iterates. By Taylor expanding, we can express $w(t+1)$ in terms of time-derivatives of $w(t)$:
\begin{equation}
    w(t+1) - w(t)  = \dot{w} + \frac{1}{2}\ddot{w} + \frac{1}{6}\dddot{w} + \cdots = \sum_{n=1}^{\infty}\frac{1}{n!} \frac{d^n w}{dt^n}
\end{equation}

Then by invoking the GD recurrence relation, we have that:

\begin{equation}\sum_{n=1}^{\infty}\frac{1}{n!} \frac{d^n w}{dt^n} = - \eta S w \label{eq:gd-bea}
\end{equation}

To solve for $\tilde{V}$, we need to express the time derivatives of $w$ in terms of $\tilde{V}$. To do so, we will use the fact that the total time derivative of any quantity $f$ along a solution is given by:
\begin{equation}
    \frac{df}{dt} = \frac{\partial f}{\partial t} + \nabla f \cdot \frac{dw}{dt}.
\end{equation}
Since our system is autonomous (the dynamics have no explicit time dependence), the partial time derivative vanishes and we have:
\begin{equation}
    \frac{df}{dt} = \nabla f \cdot \tilde{V}.
\end{equation}
This allows us to express the time derivatives of $w$ in terms of $\tilde{V}$:
\begin{align}
    \dot{w} &= \tilde{V}, \\
    \ddot{w} &= \nabla \tilde{V} \cdot \tilde{V}, \\
    \dddot{w} &= \nabla(\nabla \tilde{V} \cdot \tilde{V}) \cdot \tilde{V} = (\nabla \tilde{V})^2 \tilde{V} + \nabla^2 \tilde{V}[\tilde{V}, \tilde{V}].
\end{align}
We now invoke an ansatz: since gradient flow is linear in $w$, we expect the modified dynamics to remain linear, so we assume that $\tilde{V}(w) = -Cw$ for some constant $C$. This means that $\nabla^2 \tilde{V} = 0$, which simplifies the higher-order time derivatives immensely. In fact, we can show that:
\begin{equation}
    \frac{d^n w}{dt^n} = (\nabla \tilde{V})^{n-1} \tilde{V}
\end{equation}
for all $n \ge 1$.

We will proceed by induction. The base case $n=1$ holds trivially. For the inductive step, assume the formula holds for $n$. It then follows that:
\begin{align}
    \frac{d^{n+1} w}{dt^{n+1}} &= \nabla \left[(\nabla \tilde{V})^{n-1} \tilde{V}\right] \cdot \tilde{V} \notag \\
    &= (\nabla \tilde{V})^{n} \tilde{V} + (n-1) \nabla^2 \tilde{V}[(\nabla \tilde{V})^{n-2} \tilde{V}, \tilde{V}] \notag \\
    &= (\nabla \tilde{V})^{n} \tilde{V},
\end{align}
where the last equality uses that $\nabla^2 \tilde{V} = 0$. With our linear ansatz $\tilde{V}(w) = -Cw$, we have that $\nabla \tilde{V} = -C$. We can then see that:
\begin{equation}
    \frac{d^n w}{dt^n} = (-C)^n w.
\end{equation}
One can show that:
\begin{align}
    w(t+1) - w(t) &= \sum_{n=1}^{\infty} \frac{1}{n!} \frac{d^n w}{dt^n} \notag \\
    &= \sum_{n=1}^{\infty} \frac{(-C)^n}{n!} w \notag \\
    &= (e^{-C} - 1) w.
\end{align}
Setting this equal to the right-hand side of \cref{eq:gd-bea}, we can solve for $C$:
\begin{equation}
    (e^{-C} - 1) w = -\eta S w \implies C = -\log(1 - \eta S),
\end{equation}
which matches the easy approach.

\paragraph{The radius of convergence.} The expression $\log(1 - \eta S)$ for the modified rate reveals a fundamental limitation of the perturbative approach: the series expansion only converges for $\eta S < 1$. This is not a mere technicality---it reflects a genuine breakdown of the continuous-time approximation. When $\eta S > 1$, the factor $(1 - \eta S)$ is negative. So the iterates alternate in sign. No first-order ODE can produce sign-alternating trajectories.

This is why we do not apply backward error analysis directly to $w$ when studying edge of stability dynamics. At $\eta S = 2$---the edge of stability threshold---the iterates oscillate with fixed amplitude, and no smooth ODE can interpolate them. Rod Flow is designed to handle this regime where direct backward error analysis breaks down.

\subsection{Derivation of the $\bar{w}$ Equation}

We now apply backward error analysis to derive the $\bar{w}$ dynamics. Starting from \cref{eq:wbar_diff}:
\begin{equation}
    \bar{w}_{t+1} - \bar{w}_t = -\frac{\eta}{2}\bigl[\nabla L(\bar{w}_t + \delta_t) + \nabla L(\bar{w}_t - \delta_t)\bigr] \eqqcolon V_{\bar{w}}(\bar{w}_t, \delta_t).
\end{equation}
 Applying the backward error analysis formula (treating $\delta$ as fixed for the $\bar{w}$ equation), we have that:
\begin{equation}
    \frac{d\bar{w}}{dt} = V_{\bar{w}} - \frac{1}{2} \nabla_{\bar{w}} V_{\bar{w}} \cdot V_{\bar{w}} + O(\eta^3).
\end{equation}
We can find that the Jacobian of $V_{\bar{w}}$ is:
\begin{equation}
    \nabla_{\bar{w}} V_{\bar{w}} = -\frac{\eta}{2}\bigl[\nabla^2 L(\bar{w} + \delta) + \nabla^2 L(\bar{w} - \delta)\bigr].
\end{equation}
We then obtain the modified vector field:

\begin{align}
   \tilde{V}_{\bar{w}} &= -\frac{1}{2} \nabla_{\bar{w}} V_{\bar{w}} \cdot V_{\bar{w}} \notag \\
    &= -\frac{1}{2} \cdot \left(-\frac{\eta}{2}\right) \bigl[\nabla^2 L(\bar{w}+\delta) + \nabla^2 L(\bar{w}-\delta)\bigr] \cdot \left(-\frac{\eta}{2}\right)\bigl[\nabla L(\bar{w}+\delta) + \nabla L(\bar{w}-\delta)\bigr] \notag \\
    &= -\frac{\eta^2}{8}\bigl[\nabla^2 L(\bar{w}+\delta) + \nabla^2 L(\bar{w}-\delta)\bigr]\bigl[\nabla L(\bar{w}+\delta) + \nabla L(\bar{w}-\delta)\bigr].
\end{align}
This yields \cref{eq:wbar_ode}.

\subsection{Why No Backward Error Correction for $\Sigma$}

The $\Sigma$ equation contains a decay term $-2\Sigma$, giving it a characteristic timescale of $O(1)$. In contrast, $\bar{w}$ evolves at rate $O(\eta)$. When $\eta \ll 1$, this timescale separation means that $\Sigma$ equilibrates much faster than $\bar{w}$ changes appreciably. At quasi-equilibrium, $\Sigma$ satisfies:

\begin{equation}
    \Sigma \approx \frac{\eta^2}{8}\bigl[\nabla L(\bar{w}+\delta) \otimes \nabla L(\bar{w}+\delta) + \nabla L(\bar{w}-\delta) \otimes \nabla L(\bar{w}-\delta)\bigr].
\end{equation}

We omit the backward error correction for $\Sigma$ for two reasons. First, since $\bar{w}$ changes little over one time step, the quasi-equilibrium value of $\Sigma$ (toward which it rapidly relaxes) also changes little. Second, any correction to $\Sigma$ would be proportional to its time derivative, which is near zero when $\Sigma$ is close to quasi-equilibrium. Together, these observations imply that the backward error correction for $\Sigma$ has negligible effect.

\section{Fixed Point Analysis}
\label{app:fixedpoints}

\subsection{Quartic Potential}

Consider an even quartic potential in one dimension:
\begin{equation}
    L(w) = \frac{S}{2}w^2 + \frac{Q}{4}w^4\,,
\end{equation}

We will now perform a more thorough analysis of the quartic potential. We will require that the quadratic coefficient $S$ is greater than zero as we did in \cref{sec:theory}. But we will now consider both positive and negative quartic coefficients $Q$. We will consider four cases: $S <2/\eta$ or $S > 2/\eta$, crossed with $Q < 0$ or $Q > 0$.

\subsubsection{Deriving the $\Sigma$ Dynamics}

 The gradient is given as $\nabla L(w) = Sw + Qw^3$. At $\bar{w} = 0$ with $\delta = \sqrt{\Sigma}$, the gradients at the rod endpoints are:
\begin{equation}
    \nabla L(\pm\delta) = \pm\bigl(S\sqrt{\Sigma} + Q\Sigma^{3/2}\bigr)\,.
\end{equation}

Since the potential is symmetric, $\frac{d\bar{w}}{dt} = 0$: the center remains at the origin. The ODE governing $\Sigma$ is:
\begin{equation}
   \frac{d\Sigma}{dt} = \Bigl(\frac{\eta^2 S^2}{2} - 2\Bigr)\Sigma + \eta^2 SQ\,\Sigma^2 + \frac{\eta^2 Q^2}{2}\Sigma^3\,. \label{eq:quartic_sigma}
\end{equation}

For convenience, define the right-hand side as $f(\Sigma)$:
\begin{equation}
    f(\Sigma) = \Bigl(\frac{\eta^2 S^2}{2} - 2\Bigr)\Sigma + \eta^2 SQ\,\Sigma^2 + \frac{\eta^2 Q^2}{2}\Sigma^3\,.
\end{equation}

\subsubsection{Fixed Points}

Setting $f(\Sigma) = 0$, we find that $\Sigma^* = 0$ is always a fixed point. Factoring out $\Sigma$, the non-zero fixed points satisfy:
\begin{equation}
    \frac{\eta^2 Q^2}{2}(\Sigma^*)^2 + \eta^2 SQ\,\Sigma^* + \Bigl(\frac{\eta^2 S^2}{2} - 2\Bigr) = 0\,. \label{eq:quartic_quadratic}
\end{equation}

Applying the quadratic formula, we have that:
\begin{align}
    \Sigma^*
    &= \frac{-\eta^2 SQ \pm \sqrt{\eta^4 S^2 Q^2 - 4 \cdot \frac{\eta^2 Q^2}{2}\bigl(\frac{\eta^2 S^2}{2} - 2\bigr)}}{\eta^2 Q^2}\,.
\end{align}

Let $\Delta$ denote the discriminant. The discriminant simplifies as:
\begin{align}
    \Delta &= \eta^4 S^2 Q^2 - \eta^2 Q^2 \bigl(\eta^2 S^2 - 4\bigr) \notag\\
    &= \eta^4 S^2 Q^2 - \eta^4 S^2 Q^2 + 4\eta^2 Q^2 \notag\\
    &= 4\eta^2 Q^2\,.
\end{align}

Thus $\sqrt{\Delta} = 2\eta|Q|$, and the fixed points are:
\begin{equation}
    \Sigma^* = \frac{-\eta^2 SQ \pm 2\eta|Q|}{\eta^2 Q^2} = -\frac{1}{Q} \left (S \pm \frac{2}{\eta}  \right ).
\end{equation}

Note that because $\Sigma$ represents an extent, only non-negative values are physically meaningful.

\subsubsection{Stability Analysis}

To determine stability of a fixed point, we evaluate the derivative of $f$ at that point. A fixed point is stable if $f'(\Sigma^*) < 0$ and unstable if $f'(\Sigma^*) > 0$. We have:
\begin{equation}
    f'(\Sigma) = \frac{\partial}{\partial\Sigma}\frac{d\Sigma}{dt} = \Bigl(\frac{\eta^2 S^2}{2} - 2\Bigr) + 2\eta^2 SQ\,\Sigma + \frac{3\eta^2 Q^2}{2}\Sigma^2\,.
\end{equation}

When $\Sigma^{*} = 0$, we have that:

\begin{equation}
    f'(0) = \frac{\eta^2 S^2}{2} - 2\,.
\end{equation}

So the $\Sigma^* = 0$ fixed point is stable when:

\begin{equation}
f'(0) < 0 \implies S < \frac{2}{\eta}.
\end{equation}

At a non-zero fixed point, we can simplify our expression for $f'(\Sigma^*)$ by using the fact that, by definition, every non-zero fixed point $\Sigma^{*}$ must satisfy~\cref{eq:quartic_quadratic}. Substituting the fixed point condition:
\begin{align}
    f'(\Sigma^*) &= \Bigl(\frac{\eta^2 S^2}{2} - 2\Bigr) + 2\eta^2 SQ\,\Sigma^* + \frac{3\eta^2 Q^2}{2}(\Sigma^*)^2 \\
    &= \left(-\eta^2 SQ\,\Sigma^* - \frac{\eta^2 Q^2}{2}(\Sigma^*)^2\right) + 2\eta^2 SQ\,\Sigma^* + \frac{3\eta^2 Q^2}{2}(\Sigma^*)^2 \\
    &= \eta^2 SQ\,\Sigma^* + \eta^2 Q^2 (\Sigma^*)^2 \\
    &= \eta^2 Q^2 \Sigma^*\, \left[\frac{S}{Q}  + \Sigma^*\right]\,.
\end{align}

Define:

$$\Sigma^*_\pm = - \frac{1}{Q} \left (S \pm \frac{2}{\eta} \right) $$

We want to solve for when $\Sigma^*_\pm$ are physically significant (greater than zero) and also for when they are stable.

Recalling that $S > 0$, we have that:

\begin{equation}
\Sigma^*_+ > 0 \implies Q < 0.
\end{equation}

And that:

\begin{equation}
\Sigma^*_- > 0 \implies  S > \frac{2}{\eta} \text{  and  } Q < 0  \quad \text{ OR } \quad S < \frac{2}{\eta} \text{  and  } Q > 0.
\end{equation}

And for their stability we have that:

\begin{align}
f'(\Sigma^*_\pm) &= \eta^2 Q^2 \Sigma^*_\pm \left[ \frac{S}{Q} + \Sigma^*_\pm \right] \notag\\
&= \eta^2 Q^2 \left[ - \frac{1}{Q} \left (S \pm \frac{2}{\eta} \right) \right] \left[ \frac{S}{Q} - \frac{1}{Q} \left (S \pm \frac{2}{\eta} \right) \right] \notag \\
&= \eta^2 Q^2 \left[ - \frac{1}{Q} \left (S \pm \frac{2}{\eta} \right) \right] \left[ \mp \frac{2}{\eta Q} \right] \notag\\
&= \pm 2\eta \left( S \pm \frac{2}{\eta} \right) \notag \\
&= 2 \eta \left ( \frac{2}{\eta} \pm S \right)
\end{align}

For a fixed point to be stable, we require that $f'(\Sigma^*) < 0$.

For $\Sigma^*_+$, the derivative is $2\eta S + 4$. Since $S, \eta > 0$, this is always positive. So $\Sigma^*_+$ is always unstable.

For $\Sigma^*_-$, the derivative is $-2\eta S + 4$. Stability requires that:
$$f'(\Sigma_-^*) =-2\eta S + 4 < 0 \implies S > 2/\eta$$

Combining this with the existence condition for $\Sigma^*_- > 0$, we find that the only stable non-zero fixed point occurs when $S > 2/\eta$ and $Q < 0$.

\subsubsection{Case Analysis}

A stable non-zero fixed point only exists when $S > 2/\eta$ and $Q < 0$. We can understand this by considering all four cases and their physical meaning.

\paragraph{Case 1:} $S < 2/\eta$ and $Q > 0$.

The sharpness at the origin is below the critical threshold $2/\eta$, so the origin is stable and oscillations decay to zero. The positive quartic term means that the potential becomes steeper away from the origin. The unstable fixed point $\Sigma_-^*$ acts as a \emph{separatrix}: if $\Sigma$ is initialized above this threshold, it grows without bound. If initialized below it, it decays to zero.

\paragraph{Case 2:} $S > 2/\eta$ and $Q > 0$.

The sharpness exceeds $2/\eta$ at the origin, making it an unstable fixed point. The positive quartic term means the potential becomes even steeper away from the origin. Within this model, oscillations grow without bound. In practice, higher-order terms would eventually provide stabilization at larger amplitudes.

\paragraph{Case 3:} $S < 2/\eta$ and $Q < 0$.

The origin is stable because $S < 2/\eta$. The negative quartic term means the potential flattens away from the origin, but this does not change the qualitative behavior since the origin is already attracting. The unstable fixed point $\Sigma_+^*$ marks the boundary of the basin of attraction: trajectories initialized beyond it enter the runaway region where the potential is unbounded below. In realistic loss landscapes, positive higher-order terms prevent such runaways.

\paragraph{Case 4:} $S > 2/\eta$ and $Q < 0$.

This is the \textbf{self-stabilization} case. The sharpness at the origin exceeds $2/\eta$, so small oscillations are amplified. However, the negative quartic term means the effective curvature decreases at larger amplitudes: the rod expands until its endpoints sample flatter regions of the potential, reaching stable equilibrium at $\Sigma_-^*$. The larger fixed point $\Sigma_+^*$ is unstable and marks the boundary of the basin of attraction.

\begin{table}[h]
\centering
\begin{tabular}{|c|c|c|c|}
\hline
\textbf{Case} & \textbf{Conditions} & \textbf{Fixed Points} & \textbf{Stability} \\
\hline
1 & $S < 2/\eta$, $Q > 0$ & $0$, $\Sigma_-^*$ & Stable, Unstable \\
\hline
2 & $S > 2/\eta$, $Q > 0$ & $0$ & Unstable \\
\hline
3 & $S < 2/\eta$, $Q < 0$ & $0$, $\Sigma_+^*$ & Stable, Unstable \\
\hline
4 & $S > 2/\eta$, $Q < 0$ & $0$, $\Sigma_-^*$, $\Sigma_+^*$ & Unstable, Stable, Unstable \\
\hline
\end{tabular}
\caption{Summary of fixed points and stability for the quartic potential.}
\label{tab:quartic_cases}
\end{table}

\subsubsection{Connection to Neural Network Training}

In the context of neural network optimization, Case 4 is the most relevant.

The condition $S > 2/\eta$ corresponds to the edge of stability regime. The condition $Q < 0$ is generically expected near local minima: loss surfaces typically exhibit positive curvature that decreases at larger distances from the minimum. While $Q < 0$ makes the quartic potential unbounded below, realistic loss surfaces have positive higher-order terms that prevent runaways. The quartic analysis captures the essential self-stabilization mechanism while remaining analytically tractable.

\section{Computational Implementation of Rod Flow}
\label{app:computation}

Implementing Rod Flow on realistically-sized neural networks presents two challenges, both of which involve large matrices. First, the extent $\Sigma \in \mathbb{R}^{p \times p}$ has $p^2$ entries where $p$ is the number of parameters. For typical neural networks---which often have hundreds of thousands or even millions of parameters---storing $\Sigma$ explicitly would require terabytes of memory. Second, the $\bar{w}$ dynamics involve the Hessian $\nabla^2 L \in \mathbb{R}^{p \times p}$, which poses similar storage and computational challenges.

This appendix describes how we address both challenges: a low-rank representation for $\Sigma$ that exploits its empirically observed spectral structure, and automatic differentiation techniques that compute Hessian-vector products without ever forming the Hessian explicitly.

\subsection{Low-Rank Representation of $\Sigma$}
\label{app:low-rank-sigma}

We can understand why $\Sigma$ should be expected to be low-rank by examining the structure of its ODE. The timescale of $\Sigma$ relaxing to equilibrium is much faster than the timescale at which $\bar{w}$ changes appreciably. This means we can approximate $\Sigma$ at any given time by its steady-state value:
\begin{equation}
    \Sigma \approx \frac{\eta^2}{8} \left( \nabla L_+ \otimes \nabla L_+ + \nabla L_- \otimes \nabla L_- \right)
\end{equation}
The steady state is at most rank-2. However, in practice the gradient direction does not change much in a single time step, so $\Sigma$ is effectively rank-1 (which accords with its discrete-time counterpart $\delta_t \otimes \delta_t$).

\begin{figure}[t]
\centering
\begin{minipage}[c]{0.6\columnwidth}
\centering
\includegraphics[width=\linewidth]{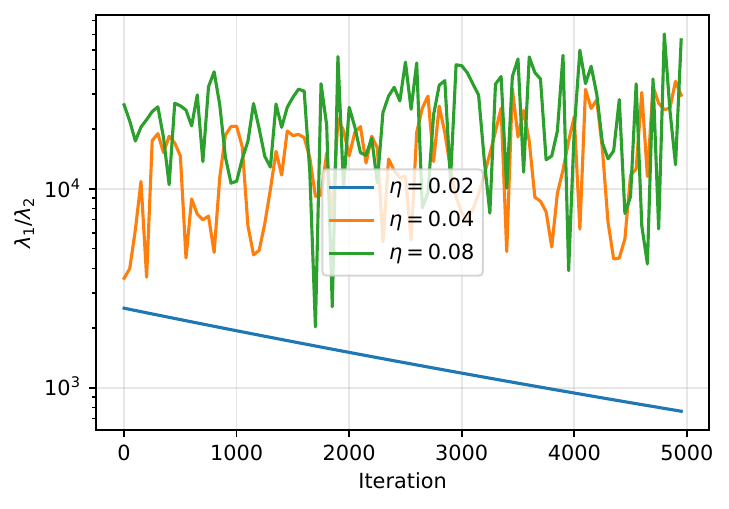}
\end{minipage}
\hfill
\begin{minipage}[c]{0.35\columnwidth}
\centering
\begin{tabular}{ccc}
\toprule
$\eta$ & Mean & Min \\
\midrule
0.02 & 1442.6 & 763.6 \\
0.04 & 14711.3 & 3555.0 \\
0.08 & 24205.7 & 2027.2 \\
\bottomrule
\end{tabular}
\end{minipage}
\caption{\textbf{Eigenvalue ratio of $\Sigma$.} Ratio of the largest to second largest eigenvalue of $\Sigma$ across learning rates, confirming that $\Sigma$ remains approximately rank-1. \textbf{Left:} Evolution during training. \textbf{Right:} Summary statistics.}
\label{fig:sigma_eigenvalue_ratio}
\end{figure}

Because $\Sigma$ is low-rank, rather than storing the entire matrix, we can store only its top few eigenvectors and eigenvalues. Specifically, we maintain $\Sigma$ in factored form:
\begin{equation}
    \Sigma = V \Lambda V^\top\,,
\end{equation}
where $V \in \mathbb{R}^{p \times r}$ is a matrix whose orthonormal columns represent the top $r$ eigenvectors of $\Sigma$, and $\Lambda = \mathrm{diag}(\lambda_1, \ldots, \lambda_r) \in \mathbb{R}^{r \times r}$ is a diagonal matrix of the corresponding eigenvalues. This reduces storage from $O(p^2)$ to $O(pr)$. We use $r = 3$ in all our experiments.

The half-difference $\delta$ is defined as the principal eigenvector of $\Sigma$ scaled by the square root of the corresponding eigenvalue . Since $\Lambda$ is diagonal, the largest eigenvalue $\lambda_1$ is simply the largest diagonal entry, and the corresponding eigenvector in the $V$ basis. Thus:
\begin{equation}
    \delta = \sqrt{\lambda_1} \cdot v_1\,,
\end{equation}
where $v_1$ is the first column of $V$.

It is not enough to store $\Sigma$. We must also update it at each time step. This presents a challenge because the update is not necessarily confined to the same low-dimensional subspace as $\Sigma$ itself. Our update proceeds in five steps.

\textbf{Step 1: Decay.} Apply the exponential decay from the $-2\Sigma$ term in the ODE:
\begin{equation}
    \Lambda \leftarrow (1 - 2\,dt) \cdot \Lambda\,.
\end{equation}
Since $\Lambda$ is diagonal, this is simply element-wise multiplication.

\textbf{Step 2: Project gradients.} Decompose each gradient into components parallel and perpendicular to the current eigenspace:
\begin{align}
    \nabla L_+^{\parallel} &= V^\top \nabla L_+\,, \qquad \nabla L_+^{\perp} = \nabla L_+ - V \nabla L_+^{\parallel}\,, \\
    \nabla L_-^{\parallel} &= V^\top \nabla L_-\,, \qquad \nabla L_-^{\perp} = \nabla L_- - V \nabla L_-^{\parallel}\,.
\end{align}
The parallel components are $r$-dimensional vectors. The perpendicular components are $p$-dimensional but orthogonal to the current basis.

\textbf{Step 3: Augment basis.} If the perpendicular components have significant norm (above a threshold $\epsilon$), add them as new basis directions:
\begin{equation}
    V_{\text{aug}} = \begin{bmatrix} V & \frac{\nabla L_+^{\perp}}{\|\nabla L_+^{\perp}\|} & \frac{\nabla L_-^{\perp}}{\|\nabla L_-^{\perp}\|} \end{bmatrix} \in \mathbb{R}^{p \times (r+k)}\,,
\end{equation}
where $k \in \{0, 1, 2\}$ depends on how many perpendicular components exceed the threshold. We orthogonalize the new directions against each other via Gram-Schmidt.

\textbf{Step 4: Update in augmented basis.} Extend $\Lambda$ to the augmented basis by padding with zeros for the new directions, then add the outer product contributions:
\begin{equation}
    \Lambda_{\text{aug}} \leftarrow \Lambda_{\text{aug}} + \frac{\eta^2}{4}dt \left[(V_{\text{aug}}^\top \nabla L_+)(V_{\text{aug}}^\top \nabla L_+)^\top + (V_{\text{aug}}^\top \nabla L_-)(V_{\text{aug}}^\top \nabla L_-)^\top\right]\,.
\end{equation}
This produces a symmetric matrix $\Lambda_{\text{aug}} \in \mathbb{R}^{(r+k) \times (r+k)}$ that is no longer diagonal.

\textbf{Step 5: Truncate.} Compute the eigendecomposition $\Lambda_{\text{aug}} = U \tilde{\Lambda} U^\top$, where the columns of $U$ are eigenvectors and $\tilde{\Lambda}$ is diagonal with eigenvalues in decreasing order. Retaining only the top $r$ eigenpairs, we update the basis by rotating into the new eigenvector coordinates:
\begin{equation}
    V \leftarrow V_{\text{aug}} U_{:,:r}\,,
\end{equation}
where $U_{:,:r}$ denotes the first $r$ columns of $U$. The new eigenvalue matrix is simply
\begin{equation}
    \Lambda \leftarrow \mathrm{diag}(\tilde{\lambda}_1, \ldots, \tilde{\lambda}_r)\,,
\end{equation}
which is diagonal by construction. We re-orthogonalize $V$ via QR decomposition to correct for numerical drift.

Each update requires $O(pr)$ for gradient projections, $O(r^3)$ for the eigendecomposition, and $O(pr)$ for updating $V$. Since $r \ll p$, the dominant cost is $O(pr)$ per update, comparable to a single gradient computation---in stark contrast to the $O(p^3)$ cost required to evolve a full $p \times p$ matrix.

\subsection{Hessian-Vector Products via Automatic Differentiation}
\label{app:hvp}

In Rod Flow, the $\bar{w}$ dynamics include a term of the form $\nabla^2 L(w) \cdot v$, where $\nabla^2 L \in \mathbb{R}^{p \times p}$ is the Hessian of the loss and $v \in \mathbb{R}^p$ is a vector (specifically, $v = \nabla L(\bar{w}+\delta) + \nabla L(\bar{w}-\delta)$).

Computing and storing the full Hessian is prohibitively expensive: forming it requires $O(p^2)$ operations and storing it requires $O(p^2)$ entries (terabytes for reasonably-sized networks). Fortunately, we never need the full Hessian---only its action on specific vectors. These are called \textit{Hessian-vector products} (HVPs). The Hessian-vector product $\nabla^2 L(w) \cdot v$ is the directional derivative of the gradient in the direction $v$:
\begin{equation}
    \nabla^2 L(w) \cdot v = \lim_{\epsilon \to 0} \frac{\nabla L(w + \epsilon v) - \nabla L(w)}{\epsilon} = \frac{d}{d \epsilon}\bigg|_{\epsilon=0} \nabla L(w + \epsilon v)\,.
\end{equation}

To compute derivatives, modern deep learning frameworks like PyTorch and JAX implement \textit{automatic differentiation} (autodiff), a technique for computing derivatives of functions defined by computer programs. Autodiff works by decomposing a computation into elementary operations and applying the chain rule systematically. There are two modes:

The \textbf{Forward mode} computes directional derivatives $\nabla f(x) \cdot v$ by propagating tangent vectors alongside the computation. Given an input perturbation $v$, forward mode tracks how this perturbation affects each intermediate value, culminating in how it affects the output. Forward mode is efficient when the number of outputs exceeds the number of inputs, as it computes derivatives with respect to one input direction per pass.

The \textbf{Reverse mode} computes gradients $\nabla f(x)$ by first performing a \textit{forward pass} that evaluates the function while recording the computation graph, then performing a \textit{backward pass} that propagates adjoints (sensitivities) from outputs back to inputs. This is the familiar ``backpropagation'' used for training neural networks. Reverse mode is efficient when the number of inputs exceeds the number of outputs---exactly the setting of neural network training, where we have millions of parameters but a scalar loss.

One can compute Hessian-vector products efficiently by combining the two modes of autodiff. The Hessian $\nabla^2 L$ is the Jacobian of the gradient $\nabla L$. Since $\nabla L: \mathbb{R}^p \to \mathbb{R}^p$ maps many inputs to many outputs, computing its full Jacobian is expensive in either mode. However, we only need the Hessian's action on a vector $v$, which is a directional derivative of the gradient---exactly what forward mode computes efficiently.

This suggests the \textit{forward-over-reverse} approach: apply forward-mode differentiation to a reverse-mode gradient computation. Concretely:
\begin{enumerate}
    \item Define the gradient as a function $g(w) = \nabla L(w)$, computed via reverse mode.
    \item Apply forward-mode differentiation to $g$ in direction $v$, yielding $\frac{d}{d\epsilon}\big|_{\epsilon=0} g(w + \epsilon v) = \nabla^2 L(w) \cdot v$.
\end{enumerate}
In PyTorch, this is implemented using \texttt{torch.func}:
\begin{verbatim}
def hvp(loss_fn, w, v):
    grad_fn = torch.func.jacrev(loss_fn)  # reverse mode
    return torch.func.jvp(grad_fn, (w,), (v,))[1]  # forward mode
\end{verbatim}
Here \texttt{jacrev} constructs a function computing gradients via reverse mode, and \texttt{jvp} differentiates that function in direction $v$ using forward mode. This requires $O(p)$ memory and roughly twice the cost of a single gradient computation, avoiding explicit construction of the $p \times p$ Hessian.

\begin{table}[h]
\centering
\begin{tabular}{lcc}
\toprule
\textbf{Operation} & \textbf{Time} & \textbf{Memory} \\
\midrule
Gradient $\nabla L(w)$ & $O(p)$ & $O(p)$ \\
Low-rank $\Sigma$ update & $O(pr + r^3)$ & $O(pr)$ \\
Hessian-vector product $\nabla^2 L(w) \cdot v$ & $O(p)$ & $O(p)$ \\
\bottomrule
\end{tabular}
\caption{Computational costs for Rod Flow operations. All operations used in our implementation scale linearly in $p$ (treating $r$ as constant), avoiding the quadratic costs of explicit matrix representations.}
\label{tab:computational-costs}
\end{table}

In total, each Rod Flow time step requires:
\begin{itemize}
    \item Two gradient evaluations: $\nabla L(\bar{w} + \delta)$ and $\nabla L(\bar{w} - \delta)$
    \item Two Hessian-vector products for the $\bar{w}$ update
    \item One low-rank $\Sigma$ update
\end{itemize}

The total cost per substep is $O(p)$, making Rod Flow practical for networks with millions of parameters. This is comparable to the cost of standard gradient descent, with only a constant factor overhead (approximately 4--6$\times$ due to the additional gradient and HVP computations).
\section{Central Flow}
\label{app:centralflow}

This appendix provides a self-contained introduction to Central Flow \citep{Cohen+25CentralFlow}. We explain the key ideas, mathematical formulation, and computational implementation of Central Flow, then show how it relates to Rod Flow and discuss the tradeoffs between the two approaches.

\subsection{Overview and Motivation}

The core idea behind Central Flow is that while the exact oscillatory dynamics of gradient descent are difficult to analyze, the \emph{time-averaged} trajectory is much more tractable. Central Flow derives an ODE for this smoothed trajectory.

Central Flow tracks two quantities:
\begin{itemize}
    \item $\bar{w}$: the center of the oscillations (the time-averaged iterate)
    \item $\Sigma$: a covariance matrix representing the spread of oscillations around the center
\end{itemize}

The original paper intentionally leaves the precise definition of ``time-averaged'' somewhat flexible.

The crucial insight, due to \citet{DamNicLee23SelfStab}, is that due to non-linearities in the loss (which can be seen by Taylor expanding the loss to third order---one order higher than the quadratic expansion common in traditional optimization), oscillations along the top Hessian eigenvector trigger a reduction of the top Hessian eigenvalue. This explains why gradient descent self-stabilizes at the edge of stability rather than diverging.

To see this, suppose gradient descent oscillates around a center $\bar{w}$ along the top eigenvector of the Hessian $u$, so the current iterate is $w = \bar{w} + xu$ for some displacement $x$. Taylor expanding the gradient:
\begin{align}
    \nabla L(\bar{w} + xu) &= \nabla L(\bar{w}) + \nabla^2L(\bar{w}) (xu) + \frac{1}{2} \nabla^3L(\bar{w})[xu, xu] + O(x^3) \\
    &= \nabla L(\bar{w}) + xS(\bar{w})u + \tfrac{1}{2}x^2 \nabla S(\bar{w}) + O(x^3)
\end{align}
where $S(\bar{w})$ is the sharpness at $\bar{w}$. The second term ($xSu$) causes the back-and-forth oscillations predicted by classical theory. The third term ($\tfrac{1}{2}x^2 \nabla S$) is the key: a gradient step implicitly takes a step in the $-\nabla S$ direction, reducing the sharpness. The strength of this sharpness reduction is proportional to $x^2$---the variance of the oscillations.

The result is self-stabilization. When the sharpness exceeds $2/\eta$, the oscillations grow. The growing oscillations then act to reduce the sharpness. Due to the reduced sharpness, the oscillations start to shrink. The shrinking oscillations allow the iterates to approach the minimum, which causes sharpness to rise again. The result is that the sharpness hovers around $2/\eta$.

\subsection{The Central Flow Equations}
\subsubsection{Dynamics of the Center}

The center $\bar{w}$ evolves according to:
\begin{equation}
    \frac{d\bar{w}}{dt} = -\eta \nabla L(\bar{w}) - \frac{\eta}{2}\Tr(\Sigma \nabla^3 L(\bar{w})),
    \label{eq:central-wbar}
\end{equation}
The first term is standard gradient flow. The second term captures the sharpness-reducing effect of oscillations.

When oscillations are concentrated along the top eigenvector, this simplifies to:
\begin{equation}
    \frac{d\bar{w}}{dt} \approx -\eta \nabla L(\bar{w}) - \frac{\eta}{2}\Tr(\Sigma) \nabla S(\bar{w}).
\end{equation}
The center moves down the loss gradient \emph{and} down the sharpness gradient with the latter weighted by $\Tr(\Sigma)$---the total oscillation variance.

\subsubsection{The Covariance Matrix}

How does one determine the covariance matrix $\Sigma$? In Central Flow's model, two properties constrain the oscillations: (1) oscillations occur only at the edge of stability, and (2) oscillations are confined to sharp directions of the Hessian. Together, these imply that $\Sigma$ has rank at most $k$, where $k$ is the number of sharp eigenvalues.

Central Flow restricts attention to the \emph{critical subspace}—the span of Hessian eigenvectors with eigenvalues near the instability threshold $2/\eta$. Let $U \in \R^{p \times k}$ denote the matrix whose columns are the top $k$ eigenvectors of $\nabla^2 L(\bar{w})$, with corresponding eigenvalues $\lambda_1, \ldots, \lambda_k$.

The covariance is parameterized within this subspace as $\Sigma = U \Omega U^\top$, where $\Omega \in \R^{k \times k}$ is positive semidefinite. Here $\Omega$ captures the covariance of oscillations expressed in the basis of critical eigenvectors. The conditions below are stated in terms of $\Omega$ rather than $\Sigma$, emphasizing that Central Flow operates on this reduced $k \times k$ representation rather than the full $p \times p$ covariance. This reduces the problem from tracking $O(p^2)$ parameters to $O(k^2)$, where typically $k \ll p$.

\subsubsection{The Semidefinite Complementarity Problem}

Central Flow determines the covariance $\Omega$ at each instant by solving the following \emph{semidefinite complementarity problem} (SDCP):
\begin{equation}
    \boxed{\Omega \succeq 0, \qquad \Lambda + \mathcal{S}[\Omega] \succeq 0, \qquad \langle \Omega, \Lambda + \mathcal{S}[\Omega] \rangle = 0}
    \label{eq:sdcp}
\end{equation}

$\Lambda$ is the \textit{stability margin}. It measures how far each eigenvalue in the critical subspace lies from the instability threshold $2/\eta$. The larger an eigenvalue of $\Lambda$, the farther that eigendirection is from the edge of stability. The margin is defined as:

\begin{equation}
\Lambda = U^\top\left(\frac{2}{\eta} I - \nabla^2 L\right)U \in \R^{k \times k}.
\end{equation}

If the columns of $U$ are eigenvectors of $\nabla^2 L$ with eigenvalues $\lambda_1, \ldots, \lambda_k$, then:

\begin{equation}
    \Lambda = \text{diag}\left(\frac{2}{\eta} - \lambda_1, \ldots, \frac{2}{\eta} - \lambda_k\right).
\end{equation}

The matrix $\Lambda$ is positive semidefinite if and only if every tracked eigenvalue of the Hessian satisfies $\lambda_i \leq 2/\eta$. Thus $\Lambda \succeq 0$ precisely encodes the requirement that all eigenvalues remain at or below the stability threshold.

$\mathcal{S}$ is the \textit{sharpness reduction operator}. The linear operator $\mathcal{S}: \R^{k \times k} \to \R^{k \times k}$ captures how oscillations reduce sharpness (and thereby increase the stability margin).

The operator $\mathcal{S}$ is closely related to the third derivative $\nabla^3 L$. Both take oscillation information as input and produce sharpness changes as output. The difference lies in their domains: $\nabla^3 L$ is a tensor acting on oscillations in the full parameter space while $\mathcal{S}$ is the restriction of this action to the critical subspace.

SDCPs arise naturally whenever inequality constraints are present. The simplest example is a ball bouncing on a floor. Let $z$ denote the ball's height and $N$ denote the normal force from the floor, with upward defined as positive for both variables. Each variable is individually constrained to be non-negative: the ball must be above the ground ($z \geq 0$) and the normal force can only push, not pull ($N \geq 0$). Furthermore, these quantities satisfy a complementarity condition: $zN = 0$. Either the ball is on the ground ($z = 0$) and the floor may exert a normal force ($N \geq 0$), or the ball is in the air ($z > 0$) and the normal force must vanish ($N = 0$). A nonzero normal force while the ball is airborne is impossible. The three conditions $z \geq 0$, $N \geq 0$, and $zN = 0$ together constitute an SDCP.

For Central Flow's SDCP: instead of requiring $z \geq 0$, we require $\Lambda + \mathcal{S}[\Omega] \succeq 0$ (the margin, after accounting for sharpness reduction from oscillations, must be positive semidefinite). Instead of $N \geq 0$, we require $\Omega \succeq 0$ (the oscillation covariance must be positive semidefinite). And instead of the scalar product $zN = 0$, we require the matrix inner product $\langle \Omega, \Lambda + \mathcal{S}[\Omega] \rangle = 0$. The complementarity condition enforces that oscillations and positive margin cannot coexist in the same direction—just as height and normal force cannot both be positive for the bouncing ball. Oscillations only occur in sharp directions of the Hessian.

Does Central Flow's SDCP have a unique solution? It does. When $\mathcal{S}$ is positive semidefinite (which holds in practice), the SDCP is equivalent to a convex quadratic program. To see this, consider the Lagrangian relaxation of the constraint $\Lambda + \mathcal{S}[\Omega] \succeq 0$. Introduce this constraint via a penalty and form the program:
\begin{equation}
    \min_{\Omega \succeq 0} \frac{1}{2}\langle \Omega, \mathcal{S}[\Omega] \rangle + \langle \Lambda, \Omega \rangle.
    \label{eq:sdcp-qp}
\end{equation}

Since \cref{eq:sdcp-qp} is a convex quadratic program over a convex domain, it has a unique solution when $\mathcal{S} \succ 0$.

\subsection{Projected Gradient Flow Interpretation}

Central Flow admits an elegant geometric interpretation as \emph{projected gradient flow}. Consider minimizing the loss subject to a constraint on the sharpness:

\begin{equation}
    \min_{\bar{w}} L(\bar{w}) \quad \text{subject to} \quad S(\bar{w}) \leq 2/\eta,
\end{equation}

The feasible region $\mathcal{F} = \{\bar{w} : S(\bar{w}) \leq 2/\eta\}$ has the edge of stability as its boundary.

Standard gradient flow $d\bar{w}/dt = -\eta\nabla L(\bar{w})$ may exit the feasible region if following the gradient increases sharpness beyond $2/\eta$. Projected gradient flow instead follows the gradient while remaining feasible:
\begin{equation}
    \frac{d\bar{w}}{dt} = -\eta\nabla L(\bar{w}) - \mu \nabla S(\bar{w}),
\end{equation}
where $\mu \geq 0$ is a Lagrange multiplier. When $S(\bar{w}) < 2/\eta$, we have $\mu = 0$ and the dynamics reduce to standard gradient flow. When $S(\bar{w}) = 2/\eta$, the multiplier $\mu$ is chosen so that the trajectory follows the gradient while remaining on the boundary.

Comparing with \cref{eq:central-wbar}, we identify $\mu = \frac{\eta}{2}\Tr(\Sigma)$. The covariance $\Sigma$ plays the role of a Lagrange multiplier: the SDCP determines exactly how much projection is needed to maintain feasibility.

\subsection{Computational Tradeoffs}

\begin{figure}[t]
    \centering
    \includegraphics[width=\textwidth]{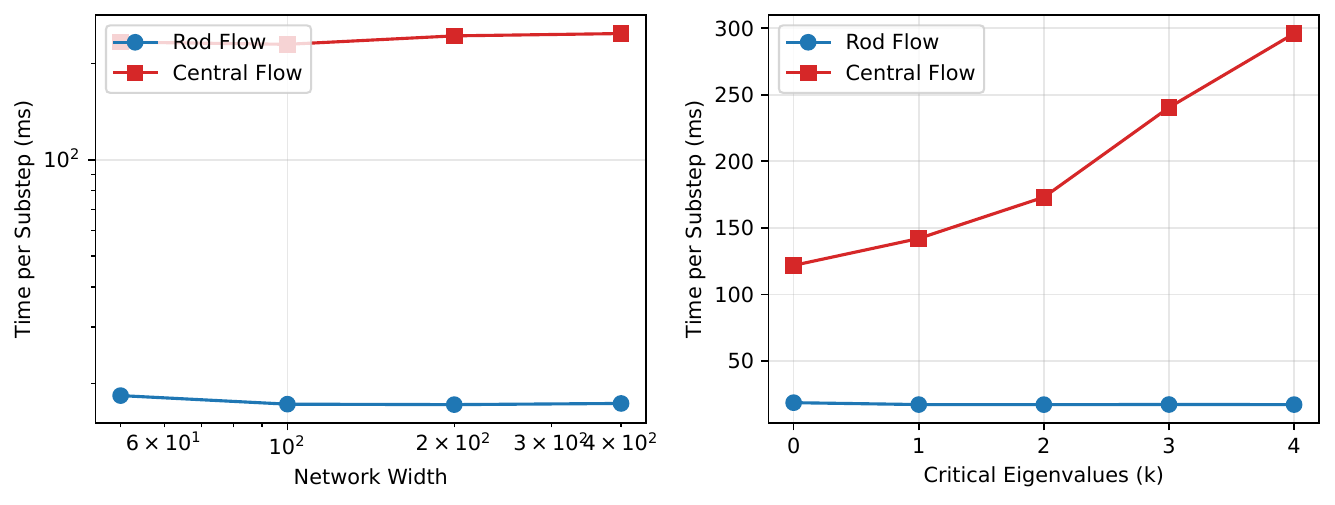}
    \caption{\textbf{Time per iteration for Rod Flow and Central Flow.} \textbf{Left:} Computation time as a function of MLP width. \textbf{Right:} Computation time as a function of the number of critical eigenvalues $k$. Rod Flow achieves substantial speedups in both regimes.}
    \label{fig:timing-comparison}
\end{figure}

\begin{table}[h]
\centering
\begin{tabular}{lcc}
\toprule
& \textbf{Rod Flow} & \textbf{Central Flow} \\
\midrule
$\Sigma$ determination & ODE integration & SDCP solve \\
Cost per step & $O(pr + r^3)$ & $O(pk + k^6)$ \\
Dependence on $k$ & None & $O(k^6)$ \\
Eigenvalue computation & Not required & Required (LOBPCG) \\
\bottomrule
\end{tabular}
\caption{Computational comparison. Here $p$ is the number of parameters, $r$ is Rod Flow's approximation rank (typically 3), and $k$ is the number of critical eigenvalues.}
\label{tab:cf-vs-rf}
\end{table}

\begin{table}[h]
\centering
\begin{tabular}{cccc}
\toprule
$k$ & Rod Flow (ms) & Central Flow (ms) & Speedup \\
\midrule
0 & 18.6 & 121.8 & 6.5$\times$ \\
1 & 17.2 & 142.0 & 8.3$\times$ \\
2 & 17.1 & 173.0 & 10.1$\times$ \\
3 & 17.2 & 240.4 & 14.0$\times$ \\
4 & 17.2 & 296.0 & 17.2$\times$ \\
\bottomrule
\end{tabular}
\caption{Timing comparison by number of critical eigenvalues $k$.}
\label{tab:timing_by_k}
\end{table}

\section{Additional Experimental Details}
\label{app:neuralnetworks}
This appendix provides additional details on the experiments performed with the 3-layer MLP, 3-layer CNN, and 2-layer Transformer. Each experiment followed a two-phase protocol.

\paragraph{Phase 1 (Warmup):} Run gradient descent until the optimizer reaches the steady-state edge of stability phase.

\paragraph{Phase 2 (Flow Comparison):} Initialize all flows (gradient flow, Rod Flow, and Central Flow) from the GD state at the end of warm-up. Then run them in lockstep with gradient descent for the remaining iterations. This enables direct comparison of how well each flow tracks the GD iterates.

\noindent We tracked the following metrics:

\begin{itemize}
    \item The loss at centers ($\bar{w}$) and edges ($\bar{w} \pm \delta$);
    \item The sharpness at centers and edges;
    \item The oscillation magnitude ($\|\delta\|$);
    \item The center discrepancy from GD ($\|\bar{w}_{\mathrm{GD}} - \bar{w}_{\mathrm{flow}}\|$);
    \item Delta alignment ($|\cos(\delta_{\mathrm{GD}}, \delta_{\mathrm{flow}})|$).
\end{itemize}

Due to the size of the weight vectors, we had to be careful about memory management. All experiments use a ``lockstep'' execution strategy where gradient descent, Gradient Flow, Rod Flow, and Central Flow run simultaneously. This avoids storing the weight history (${\sim}10$GB for long runs) at the cost of not being able to restart the flows from arbitrary points. Explicit garbage collection (\texttt{gc.collect()}) and CUDA cache clearing were performed periodically (every 20 iterations for MLP, every 500 for CNN and Transformer) to prevent memory accumulation.

\begin{figure}[t]
\centering

\begin{minipage}{\textwidth}
\centering
\small
\textbf{(a) MLP Configuration}\\[0.5em]
\begin{tabular}{ll}
\toprule
\textbf{Parameter} & \textbf{Value} \\
\midrule
Structure & Flatten $\to$ Linear(3072, 50) $\to$ SiLU $\to$ Linear(50, 50) $\to$ SiLU $\to$ Linear(50, 10) \\
Total parameters & 156,710 \\
Dataset & CIFAR-10 (first 1,000 samples), MSE loss \\
Learning rate $\eta$ & 0.025 (stability threshold $2/\eta = 80$) \\
Iterations & 26,000 total, 22,000 warmup \\
Eigenvalue computation & Warm-started LOBPCG, top 6, every 10 iterations \\
\bottomrule
\end{tabular}
\end{minipage}

\vspace{1.5em}

\begin{minipage}{\textwidth}
\centering
\small
\textbf{(b) CNN Configuration}\\[0.5em]
\begin{tabular}{ll}
\toprule
\textbf{Parameter} & \textbf{Value} \\
\midrule
\multirow{3}{*}{Structure} & Conv2d(3, 8, kernel=3, padding=1) $\to$ SiLU $\to$ AvgPool2d(2) \\
& Conv2d(8, 8, kernel=3, padding=1) $\to$ SiLU $\to$ AvgPool2d(2) \\
& Flatten $\to$ Linear($8 \times 8 \times 8$, 10) \\
Total parameters & 5,938 \\
Dataset & CIFAR-10 (first 1,000 samples), MSE loss \\
Learning rate $\eta$ & 0.05 (stability threshold $2/\eta = 40$) \\
Iterations & 24,000 total, 20,000 warmup \\
Eigenvalue computation & Scipy Lanczos, top 3, every 500 iterations (from 18,000) \\
\bottomrule
\end{tabular}
\end{minipage}

\vspace{1.5em}

\begin{minipage}{\textwidth}
\centering
\small
\textbf{(c) Transformer Configuration}\\[0.5em]
\begin{tabular}{ll}
\toprule
\textbf{Parameter} & \textbf{Value} \\
\midrule
Embeddings & Token: Embedding(30522, 16), Position: Embedding(32, 16) \\
Transformer layers & 2 layers: 2-head attention (head dim=8), MLP (16$\to$16$\to$16), LayerNorm \\
Output & Mean pooling $\to$ Linear(16, 1) \\
Total parameters & 492,273 \\
Dataset & SST-2 (first 1,000 samples), 32 tokens, MSE loss \\
Learning rate $\eta$ & 0.01 (stability threshold $2/\eta = 200$) \\
Iterations & 31,000 total, 27,000 warmup \\
Eigenvalue computation & Scipy Lanczos, top 3, every 500 iterations (from 26,000) \\
\bottomrule
\end{tabular}
\end{minipage}

\caption{Experimental configurations for neural network experiments.}
\label{fig:experiment_configs}
\end{figure}

\begin{figure}[t]
\centering
\includegraphics[width=0.9\columnwidth]{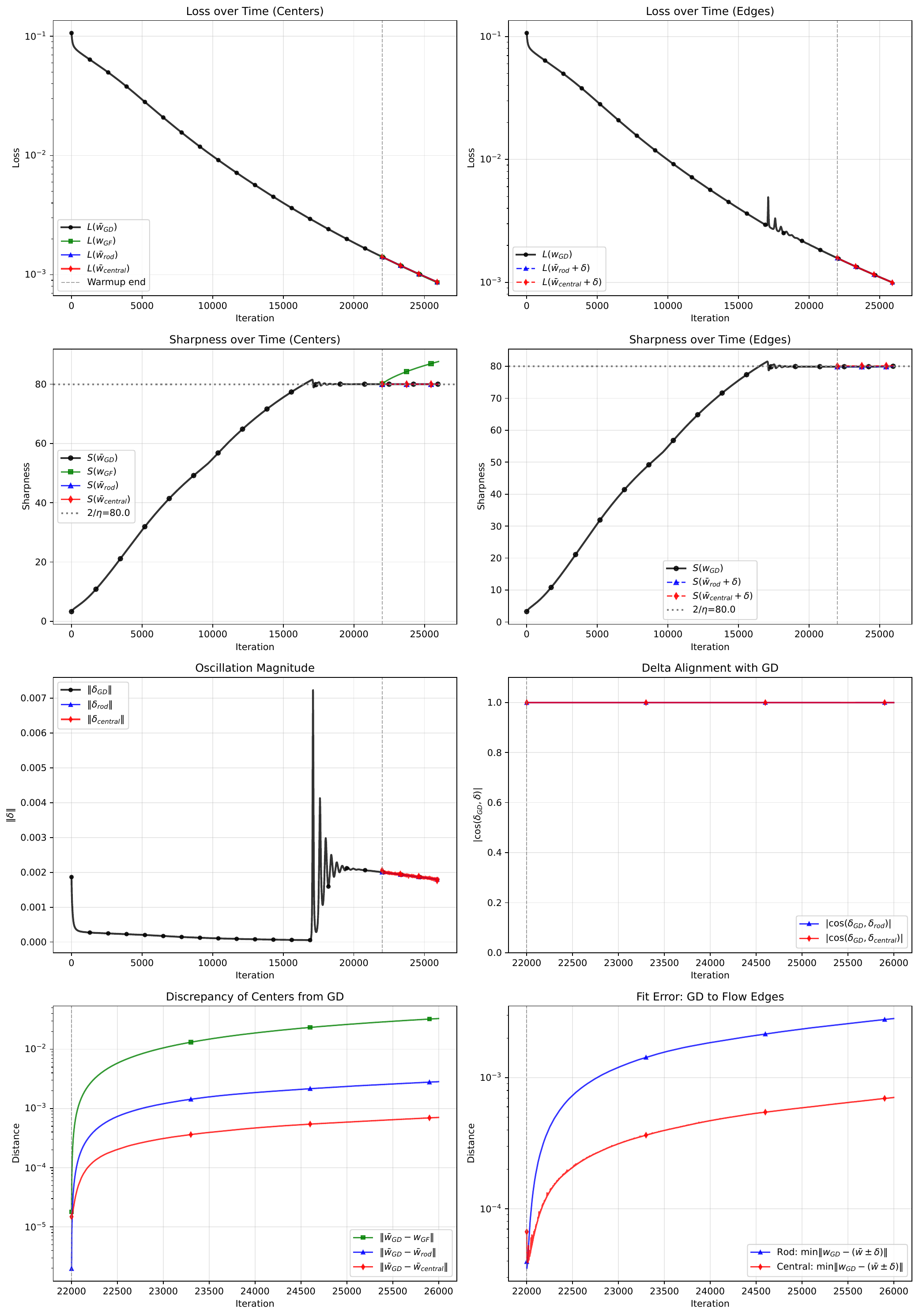}
\caption{\textbf{Flow comparison on 3-layer MLP.}}
\label{fig:mlp_flow_comparison}
\end{figure}

\begin{figure}[t]
\centering
\includegraphics[width=0.9\columnwidth]{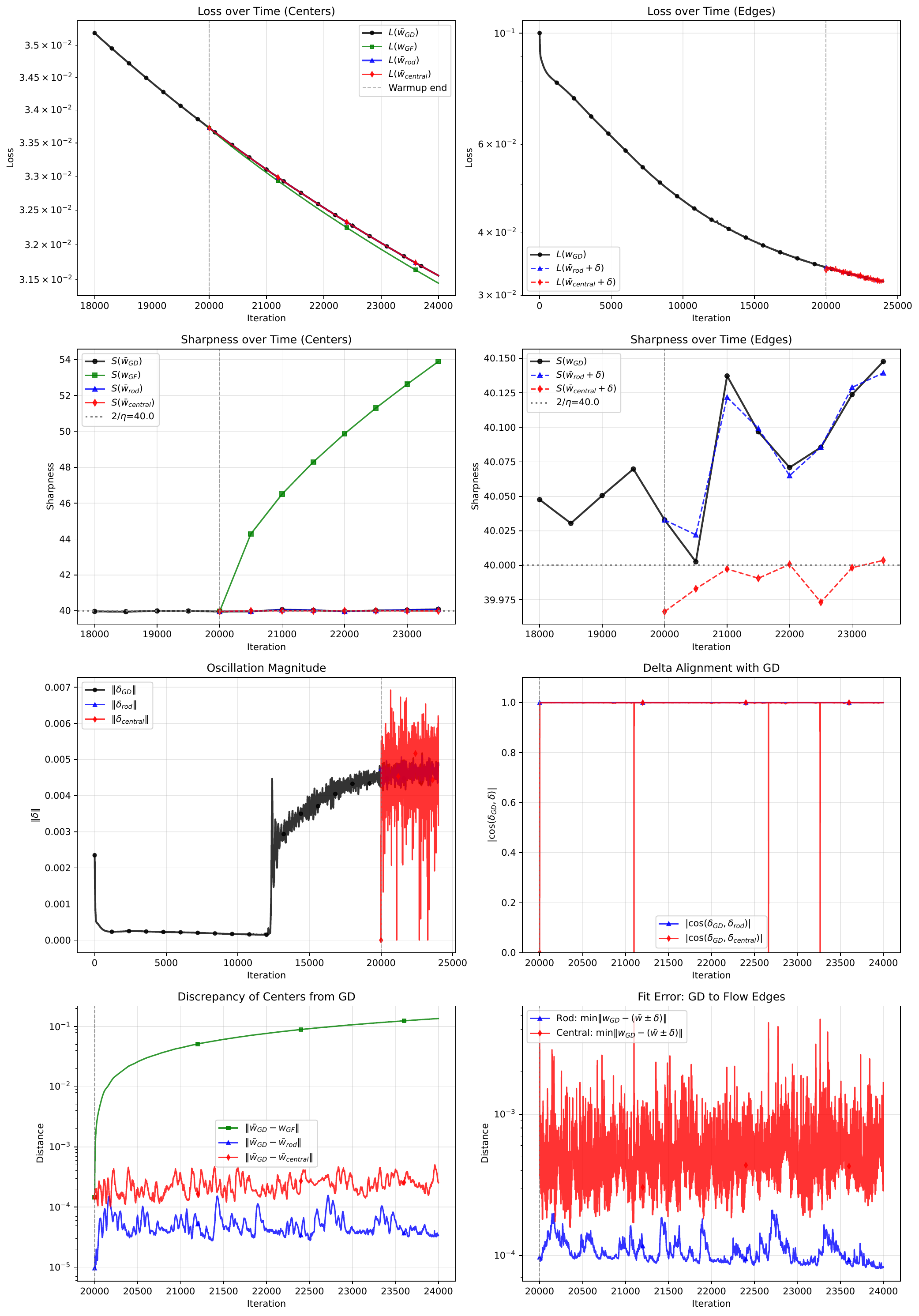}
\caption{\textbf{Flow comparison on 3-layer CNN.}}
\label{fig:cnn_flow_comparison}
\end{figure}

\begin{figure}[t]
\centering
\includegraphics[width=0.9\columnwidth]{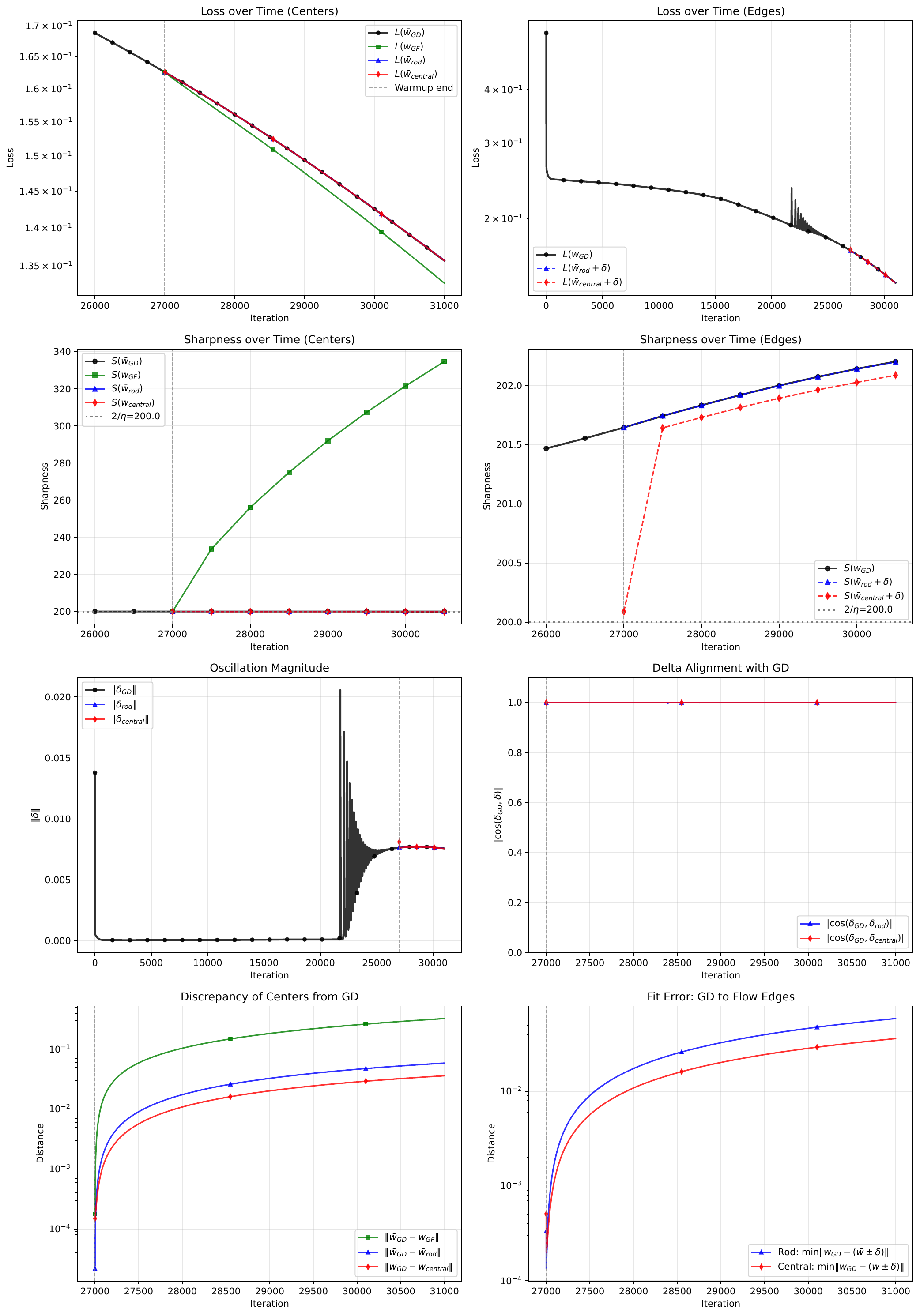}
\caption{\textbf{Flow comparison on Transformer.} }
\label{fig:transformer_flow_comparison}
\end{figure}

Results are shown in \cref{fig:mlp_flow_comparison} (MLP), \cref{fig:cnn_flow_comparison} (CNN), and \cref{fig:transformer_flow_comparison} (Transformer).

\paragraph{MLP on CIFAR-10}

The hidden dimension of the MLP was reduced from the standard 200 used in prior work to 50. We observed that Rod Flow tracks gradient descent more closely for smaller networks, likely because the oscillation dynamics are lower-dimensional and better captured by the low-rank $\Sigma$

\paragraph{CNN on CIFAR-10}

The number of filters was reduced from the standard 32 to 8, primarily for computational efficiency. Computing Hessian-vector products through convolutional layers requires backpropagating through the convolution operation twice (once for the gradient, once for the Hessian-vector product), with cost scaling quadratically in the number of filters for the dominant terms. With 8 filters, experiments complete in reasonable time while still exhibiting clear edge of stability behavior, and the smaller parameter count (5,938 vs.\ ${\sim}80,000$ with 32 filters) provides a more favorable setting. For eigenvalue computation, we used Scipy's sparse Lanczos solver instead of LOBPCG, as we found that LOBPCG occasionally failed to converge reliably for the CNN Hessian structure.

\paragraph{Transformer on SST-2}

The hidden size was reduced from a typical 64 to 16, and sequence length was capped at 32 tokens. Both choices are primarily for computational efficiency: Transformer parameter count scales as $O(d^2)$ in hidden size $d$, so reducing from 64 to 16 yields approximately $16\times$ speedup in HVP computation. Self-attention has $O(n^2)$ complexity in sequence length $n$, affecting both forward passes and HVP computation. We determined that 32 tokens is sufficient for SST-2's short sentences. The smaller learning rate compared to the MLP and CNN experiments was necessary for numerical stability given the large parameter count and sensitivity of attention mechanisms. Computing HVPs through attention layers is particularly expensive because softmax attention weights must be recomputed during both forward and backward passes, with the quadratic sequence-length dependence appearing in both the attention computation and its derivatives. Despite these costs, the Transformer experiments demonstrate that Rod Flow generalizes beyond simple architectures to modern deep learning building blocks.



\begin{table}[h]
\centering
\caption{Summary of neural network experiment configurations.}
\label{tab:experiment_config}
\begin{tabular}{lccc}
\toprule
& \textbf{MLP} & \textbf{CNN} & \textbf{Transformer} \\
\midrule
Parameters & 156,710 & 5,938 & 492,273 \\
Dataset & CIFAR-10 & CIFAR-10 & SST-2 \\
Samples & 1,000 & 1,000 & 1,000 \\
Learning rate $\eta$ & 0.025 & 0.05 & 0.01 \\
$2/\eta$ & 80 & 40 & 200 \\
Total iterations & 26,000 & 24,000 & 31,000 \\
Warmup iterations & 22,000 & 20,000 & 27,000 \\
Eigenvalue solver & LOBPCG & Scipy & Scipy \\
\bottomrule
\end{tabular}
\end{table}

\end{document}